\PassOptionsToPackage{table}{xcolor} 
\documentclass[sigconf, nonacm]{acmart}
\usepackage{tikz}
\usepackage{amsmath}
\usepackage{bm}
\usepackage{booktabs} 
\usepackage{siunitx} 
\usepackage[table]{xcolor}
\usepackage{graphicx}
\usepackage{hyperref}
\usepackage{pifont}

\newcommand{\notcheckmark}{{$\checkmark$}\textsuperscript{\textcolor{black}{\kern-0.9em{\bf---}}}}

\usepackage{tcolorbox}
\newtcolorbox{remark}{
  colback=gray!10,       
  colframe=black!75,     
  rounded corners          
}

\AtBeginDocument{%
  }

\acmConference[SACMAT '25]{SACMAT '25: ACM Symposium on Access Control Models and Technologies}{July 08--10, 2025}{Stony Brook, NY}

\begin{document}

\title{MUBox: A Critical Evaluation Framework of Deep~Machine~Unlearning}

\author{Xiang Li}
\email{xl5@fordham.edu}
\orcid{0009-0007-7467-6221}
\affiliation{%
  \institution{Fordham University}
  \city{New York City}
  \state{New York}
  \country{USA}
}

\author{Bhavani Thuraisingham}
\email{bxt043000@utdallas.edu}
\orcid{0000-0003-4653-2080}
\affiliation{%
  \institution{The University of Texas at Dallas}
  \city{Richardson}
  \state{Texas}
  \country{USA}
}

\author{Wenqi Wei}
\email{wwei23@fordham.edu}
\orcid{0000-0003-2560-4225}
\affiliation{%
  \institution{Fordham University}
  \city{New York City}
  \state{New York}
  \country{USA}
}

\begin{abstract}
Recent legal frameworks have mandated the \textit{right to be forgotten}, obligating the removal of specific data upon user requests. Machine Unlearning has emerged as a promising solution by selectively removing learned information from machine learning models. This paper presents MUBox, a comprehensive platform designed to evaluate unlearning methods in deep learning. MUBox integrates 23 advanced unlearning techniques, tested across six practical scenarios with 11 diverse evaluation metrics. It allows researchers and practitioners to (1) assess and compare the effectiveness of different machine unlearning methods across various scenarios; (2) examine the impact of current evaluation metrics on unlearning performance; and (3) conduct detailed comparative studies on machine unlearning in a unified framework. Leveraging MUBox, we systematically evaluate these unlearning methods in deep learning and uncover a set of key insights: (a) Even state-of-the-art unlearning methods, including those published in top-tier venues and winners of unlearning competitions, demonstrate inconsistent effectiveness across diverse scenarios. Prior research has predominantly focused on simplified settings, such as random forgetting and class-wise unlearning, highlighting the need for broader evaluations across more complex and realistic unlearning tasks. (b) Assessing unlearning performance remains a non-trivial problem, as no single evaluation metric can comprehensively capture the effectiveness, efficiency, and preservation of model utility. Our findings emphasize the necessity of employing multiple metrics to achieve a balanced and holistic assessment of unlearning methods. (c) In the context of depoisoning-removing the adverse effects of poisoned data-our evaluation reveals significant variability in the effectiveness of existing approaches, which is highly dependent on the specific type of poisoning attack.
We believe MUBox will serve as a valuable benchmark, advancing research in machine unlearning and highlighting areas for future improvement. Codes are available at \url{https://github.com/Jessegator/MUBox}.
\end{abstract}


\maketitle

\section{Introduction}
Machine learning (ML) models are trained on vast amounts of data that often contain sensitive, private, or copyrighted information. These models can inadvertently pose privacy risks by retaining and potentially exposing sensitive information from their training data~\cite{song2021systematic,salem2020updates,shokri2017membership}.  

Moreover, recent legislation such as the General Data Protection Regulation (GDPR) in the European Union~\cite{mantelero2013eu}, the California Consumer Privacy Act (CCPA)~\cite{harding2019understanding} in the USA, the Personal Information Protection and Electronic Documents Act (PIPEDA) in Canada~\cite{canada_law}, and the Personal Information Protection Law of China~\cite{chen2021understanding} mandate service providers to comply with users' \textit{right to be forgotten}~\cite{shastri2019seven} upon receiving the data removal requests.

In response to increasing data privacy regulations, Machine Unlearning has emerged as a promising solution to selectively remove learned knowledge from ML models. This process enables ML models to forget specific data upon request, such as copyrighted data~\cite{li2024machine} and sensitive personally identifiable information (PII) data~\cite{layne2024}. While retraining a model from scratch is a straightforward solution, it is computationally expensive and impractical for large-scale models. To address this challenge, researchers have developed various unlearning methods~\cite{guo2020certified,baumhauer2022machine,golatkar2021mixed,wu2020deltagrad,izzo2021approximate,neel2021descent,bourtoule2021machine,thudi2022unrolling,heng2024selective,warnecke2023machine,chundawat2023can,kurmanji2023towards,Mehta_2022_CVPR,GradientProjectionUnlearning,golatkar2020forgetting,golatkar2020eternal,foster2024fast,foster2024zero,chen2023boundary,jia2023model,tarun2023fast,fan2024salun,chundawat2023zero}, 
employing diverse strategies. 

Security researchers, practitioners, and service providers are confronted with a myriad of machine unlearning techniques available in the literature. However, there is a
notable lack of comprehensive, unbiased evaluations, leaving a significant gap in the quantitative understanding of the strengths and limitations of these methods.
\textit{First}, existing evaluations often employ different model architectures and datasets, making it challenging to directly compare the effectiveness of various unlearning approaches. \textit{Second}, existing evaluations are often limited to specific unlearning scenarios, such as random data forgetting, where a subset of training data from all classes is randomly selected to be unlearned. This randomness introduces bias, as certain data subsets may represent worst/best-case scenarios for unlearning, a factor not adequately addressed by current approaches. \textit{Third}, existing evaluations often use different metrics, resulting in inconsistent and potentially biased comparisons that may favor certain methods. 

To advance the research on machine unlearning, it is essential to develop a comprehensive evaluation and analysis platform for rigorous assessment. However, constructing such a platform presents several key challenges: (1) ensuring a unified comparison of various unlearning methods using the same dataset and model configurations can be difficult due to differing data and model requirements; (2) incorporating a broad range of representative unlearning methods for comprehensive evaluation demands substantial effort, given the rapid evolution and diversity of unlearning techniques; (3) providing a diverse array of metrics to evaluate different unlearning methods under various unlearning scenarios requires consideration of multiple perspectives, e.g., efficiency and effectiveness of unlearning and accuracy of the unlearned model. 
To date, no existing work~\cite{grimes2024gone,fan2024challenging,li2024machinesurvey,xu2024machine} meets all the requirements.
These studies either fail to provide practical guidance on the interaction between algorithms and implementations or overlook real-world applications and performance across diverse unlearning scenarios.

To bridge this gap, we present MUBox, a framework designed to comprehensively evaluate and compare unlearning methods while fulfilling all the aforementioned essential criteria. MUBox enables researchers and practitioners to understand the effectiveness of various unlearning methods and conduct comparative studies across different unlearning scenarios in a uniform, comprehensive, informative, and extensible manner. Our contributions are as follows:
\begin{itemize}
    \item We thoroughly review machine unlearning papers published in top security, AI, and ML venues over the past five years (2019-2024). Among 49 deep machine unlearning papers, our reproducibility study reveals that 30.6\% of them fail to include code for reproducing their experiments, and 50.0\% of the provided code lacks detailed instructions or has no instructions at all. This highlights the urgent need for a benchmarking framework to enhance reproducibility, provide clear implementation guidelines, and foster trust in machine unlearning research.
    \item We present MUBox, the \textbf{first unified evaluation platform developed for machine unlearning}, currently featuring 23 unlearning approaches in deep learning, 6 practical unlearning scenarios and 11 evaluation metrics for utility and efficacy criteria. To the best of our knowledge, \textbf{MUBox offers the most extensive collection of unlearning methods and comprehensive evaluation metrics available to date}.
    \item Our investigation offers several key insights that could drive future advancements in machine unlearning: (a) Our extensive evaluation reveals that even the state-of-the-art unlearning methods published on top-tier conferences/journals and winning the first place at Unlearning competitions lack consistent effectiveness across diverse scenarios. Notably, prior research has primarily focused on random forgetting and class-wise unlearning—the simplest scenarios. We advocate for future studies to evaluate methods across a broader range of unlearning scenarios; (b) Our evaluation indicates that designing effective metrics to assess unlearning performance remains non-trivial. No single metric can capture all aspects of unlearning, such as effectiveness, efficiency, and the preservation of model utility. 
     This necessitates the use of multiple evaluation metrics to provide a more comprehensive and balanced assessment; and (c) Our evaluation of various unlearning methods for depoisoning—removing the adverse effects of poisoned data on a model—reveals that not all existing approaches are equally effective for this task. The performance of depoisoning varies significantly depending on the different types of poisoning attacks.
\end{itemize}

In summary, MUBox offers a \textbf{systematic}, \textbf{comprehensive}, and \textbf{evidence-based} evaluation framework. MUBox provides the flexibility to evaluate whether the unlearned models meet their specific utility/efficacy criteria and facilitates the comparison of all available unlearning methods. We believe MUBox can serve as a valuable platform for identifying gaps, limitations, and opportunities for future machine unlearning research and deployment.

\section{Machine Unlearning: the Hype and Reality}
\label{sec:mu_stat_analysis}

\subsection{Machine Unlearning Concept}

\textbf{Machine unlearning}~\cite{cao2015towards} refers to the process of selectively removing learned knowledge from machine learning models to comply with data removal requests. Consider a training dataset $\mathcal{D} = \{z_i = (x_i, y_i)\}^n_{i=1}$ consisting of $n$ samples, where $x_i$ is the $i$-th sample and $y_i$ is the corresponding class label. Let $\mathcal{D}_f \subseteq \mathcal{D}$ represent the subset of data to be unlearned, and $\mathcal{D}_r = \mathcal{D} \backslash \mathcal{D}_f$ be the retaining dataset, i.e.,  $\mathcal{D}_r \cup \mathcal{D}_f = \mathcal{D}$ and $\mathcal{D}_r \cap \mathcal{D}_f = \emptyset$. Let $\mathcal{D}_t$ denote the test dataset, $\mathcal{A}(\cdot)$ as the learning algorithm, and $\mathcal{U}(\cdot)$ as the unlearning process. The original model, trained on the full dataset $\mathcal{D}$, is denoted as $\theta_o = \mathcal{A}(\mathcal{D}) \in \Theta$, and the corresponding unlearned model is represented as $\theta_u = \mathcal{U}(\mathcal{A}(D), \mathcal{D}, \mathcal{D}_f)$. Ideally, the optimal unlearned model $\theta_u^* = \mathcal{A}(\mathcal{D}_r)$, should act as if it were trained from scratch on the retaining dataset $\mathcal{D}_r $. Existing machine unlearning methods can be categorized into exact unlearning and approximate unlearning.

\textbf{Exact unlearning}~\cite{golatkar2020eternal}. Exact unlearning aims to completely remove the influence of forgetting data from the model, making the unlearned model statistically indistinguishable from one trained solely on the retaining dataset. Given a distribution measure $\mathcal{K}(\cdot)$, the unlearning process $\mathcal{U}(\cdot)$ achieve exact unlearning if, 
\begin{equation}
\mathcal{K}\left(\mathcal{P}(\mathcal{U}(\mathcal{A}(D), \mathcal{D}, \mathcal{D}_f)), \mathcal{P}(\mathcal{A}(\mathcal{D}_r)) \right) = 0 
\end{equation}
where $\mathcal{P}(\cdot)$ is the distribution of the model's parameters. 

\textbf{Approximate unlearning}~\cite{thudi2022unrolling}. Approximate unlearning relaxes the exact unlearning condition by requiring the learning distributions of the unlearned model and a retrained model to be approximately indistinguishable. The unlearning process  $\mathcal{U}(\cdot)$ satisfies $(\epsilon,\delta)$-unlearning if, 
\begin{equation}
\mathcal{P}(\mathcal{U}(\mathcal{A}(D), \mathcal{D}, \mathcal{D}_f)) \leq e^{\epsilon}\mathcal{P}(\mathcal{A}(\mathcal{D}_r))+\delta
\end{equation}
The objective of machine unlearning is to successfully erase the information $\mathcal{D}_f$ from the model without compromising its performance on the retained data $\mathcal{D}_r$ and the test dataset $\mathcal{D}_t$.

\subsection{The Urgency in Reproducibility}

The reproducibility of machine learning within the security community remains a significant yet challenging issue~\cite{olszewski2023get}. To better understand the current state of machine unlearning research, especially in deep learning, we conduct a comprehensive study to assess the computational reproducibility of recent literature.

\textit{Paper selection}. We collect machine unlearning papers published between 2019 and 2024 from the following sources: (1) top security conferences and journals, including IEEE S\&P, ACM CCS, USENIX Security, Euro S\&P, NDSS, IEEE T-IFS, IEEE TDSC; (2) leading AI and ML venues including NeurIPS, ICLR, ICML, COLT, ALT, AAAI, IJCAI, IEEE TNNLS, and Machine Learning; and (3) premier computer vision conferences including CVPR, ECCV, ICCV, ACM MM, and WACV.

Specifically, we select papers based on the following criteria: (1) the paper focuses on machine unlearning in deep learning, specifically addressing the image classification problem; (2) the paper creates an unlearning procedure, typically outlined in the Methodology or Experiments section; (3) the Experiments section includes a clear empirical study of the proposed unlearning approaches in deep learning, with specified evaluation metrics. After applying these criteria, we have identified 49 papers for our study. 

\begin{figure}[!t]
\centering
\includegraphics[width=0.95\columnwidth]{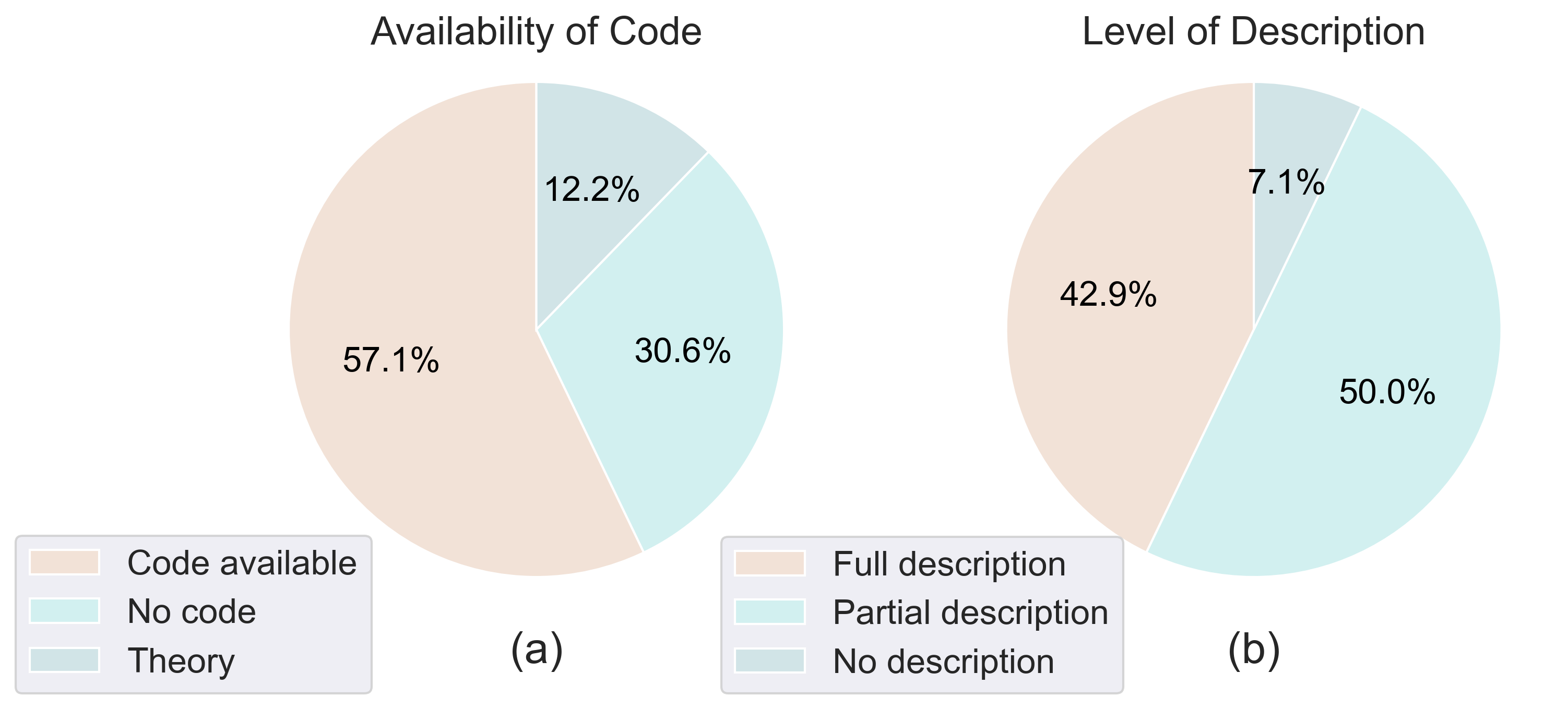}
\caption{Reproducibility analysis of 49 collected papers.}
\label{fig:statistics}
\end{figure}

\textit{Availability of code}. The unavailability of experimental code significantly hinders the ability to build upon existing research and benchmark against established methods. While many papers attempt to detail their methodologies, the complexity of novel analytical techniques and intricate system designs are often non-trivial to build from scratch based solely on the paper's description. Providing code implementations not only enhances reproducibility but also facilitates the further development of these techniques. However, as shown in \textbf{Figure~\ref{fig:statistics} (a)}, among the 49 collected papers, only 57.1\% made their code available, while 30.6\% did not provide the code necessary to reproduce their experiments, and 12.2\% only focus on theoretical analysis of machine unlearning without any code implementation.

\textit{Code documentation}. For papers whose codes are available, we further examine the quality of their README file, which is crucial for experimental reproducibility and usually comprises (1) environment setup and configuration, (2) commands to run, and (3) instructions for adjusting hyperparameters. We categorized the README files as ``Full description" if they covered all three aspects, ``Partial description" if they addressed some but not all, and ``No description" if they only included the paper title or citation information. As illustrated in \textbf{Figure~\ref{fig:statistics} (b)}, 42.9\% of those papers with codes provide detailed instructions, though not always for every experiment in their paper. By comparison, 50.0\% offer only partial guidelines, and 7.1\% give no instruction for their reproducibility at all. We also find that some of the code available requires further coding or debugging to be functional.  
It worth noting that, the NeuraIPS Machine Unlearning Competition 2023\footnote{\url{https://unlearning-challenge.github.io/}} has promoted the available code in the Unlearning community. However, in Section~\ref{sec:evaluation}, we find that these unlearning methods still lack consistent performance across different unlearning scenarios.

Additionally, We also notice significant variations in the model architectures, datasets, dataset splits, and evaluation metrics used across different studies,  highlighting the need for building a standardized unlearning benchmark.

\begin{remark}
\textbf{Remark 1.} \textit{Current machine unlearning research suffers from limited reproducibility and inconsistent evaluations across models, datasets, unlearning scenarios, and evaluation metrics, risking biased assessments and limited understanding of their performance.}
\end{remark}

\section{MUBox System Design and Implementation}
\label{sec:system}

\begin{figure}[!t]
\centering
\includegraphics[width=0.95\columnwidth]{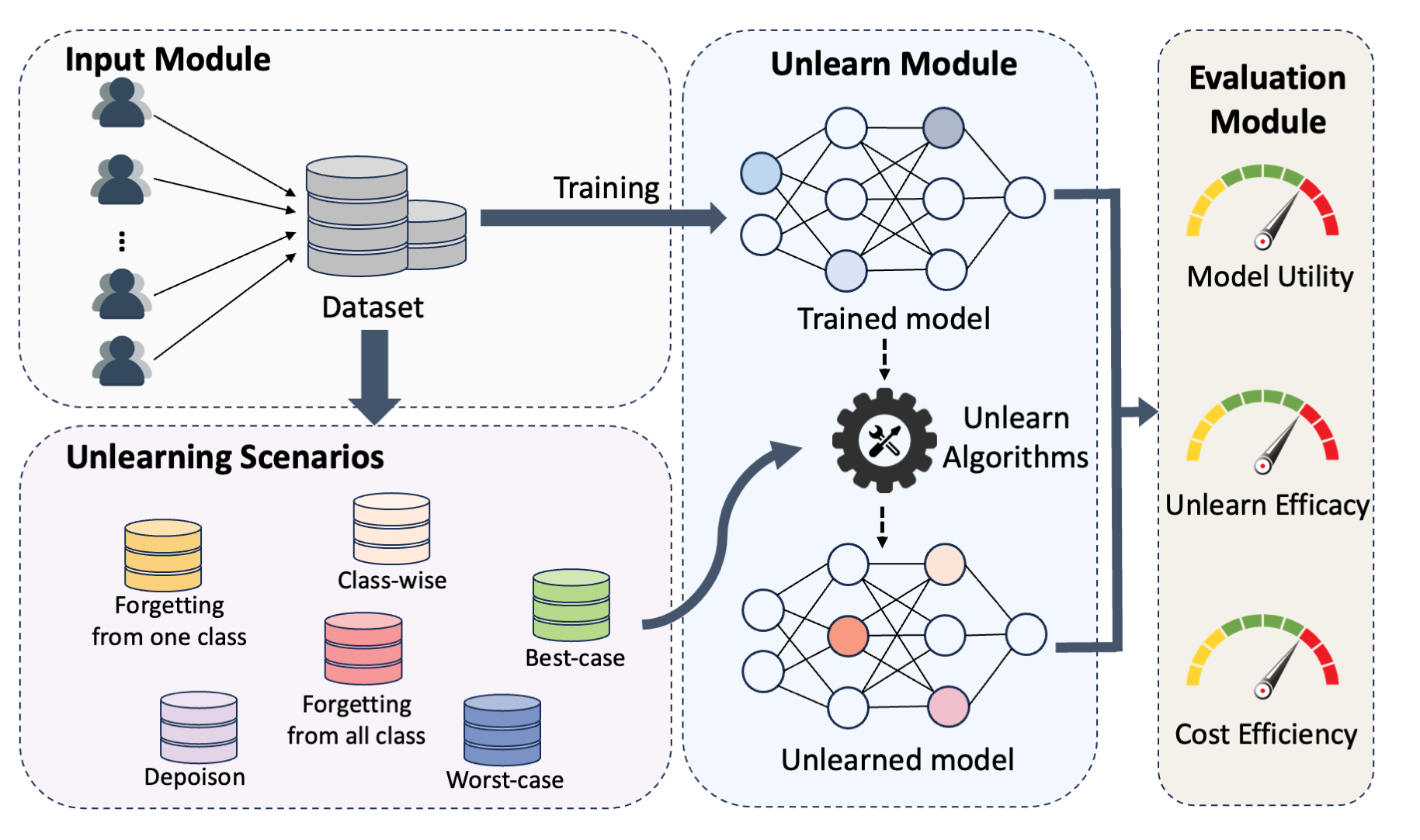}
\caption{The system overview of MUBox.}
\label{fig:overview}
\end{figure}

\subsection{System Design}
To promote the reproducibility, MUBox is the first unified evaluation platform that offers a comprehensive and systematic analysis of state-of-the-art machine unlearning algorithms across different unlearning methods under diverse unlearning scenarios. 

The system overview is illustrated in \textbf{Figure~\ref{fig:overview}}, comprising three major components. \textbf{Input Module} comprises 6 practical unlearning application scenarios (see Sec.~\ref{sec:unlean_scenarios}), which divide the entire training dataset into retaining dataset, forgetting dataset, and sub-retaining dataset.  The sub-retaining dataset is a subset of the retaining dataset required by some unlearning methods for preserving model utility. \textbf{Unlearn Module} includes a set of unlearning methods capable of generating unlearned models from the originally trained model and the data designated for unlearning. 

\textbf{Evaluation Module} is equipped with model utility metrics, unlearning efficacy metrics, and cost efficiency metrics. 

\subsection{System Implementation}

\textbf{Datasets and network architectures}.
Our choice of model architectures and datasets is based on what is commonly used in the machine unlearning community. For the discussion in this paper, we utilize two widely recognized benchmark datasets, CIFAR-10~\cite{krizhevsky2010convolutional} and TinyImageNet~\cite{deng2009imagenet}. The CIFAR-10 dataset contains 50,000 training images and 10,000 test images from 10 classes. Each image has a resolution of 3$\times$32$\times$32. The TinyImageNet dataset contains 100,000 training images, 10,000 validation images, and 10,000 test images from 200 classes. Each image has dimensions of 3$\times$64$\times$64. We select the first 50 classes of the TinyImageNet dataset. We train a ResNet-18~\cite{he2016deep} on CIFAR10 with 93.24\% test accuracy and MobileViT~\cite{mehta2022mobilevit} on TinyImageNet with 59.88\% test accuracy.

\textbf{Implementations.} So far, we have implemented a total of 23 unlearning methods, including two exact unlearning methods: Retrain and SISA~\cite{bourtoule2021machine}, and 21 approximate unlearning approaches:  Unrolling~\cite{thudi2022unrolling}, Amnesiac~\cite{heng2024selective}, First-order~\cite{warnecke2023machine}, Second-order~\cite{warnecke2023machine}, Bad-T~\cite{chundawat2023can}, SCRUB~\cite{kurmanji2023towards},
L-CODEC~\cite{Mehta_2022_CVPR}, PGU~\cite{GradientProjectionUnlearning}, Fisher~\cite{golatkar2020eternal}, NTK~\cite{golatkar2020forgetting}, SSD~\cite{foster2024fast}, Boundary-S~\cite{chen2023boundary}, Boundary-E~\cite{chen2023boundary}, $l_1$-sparsity~\cite{jia2023model}, SalUn~\cite{fan2024salun}, FCS~\cite{FCS}, MSG~\cite{MSG}, CT~\cite{CT}, NIU~\cite{NIU}, UNSIR~\cite{tarun2023fast}, and GKT~\cite{chundawat2023zero}.

\textbf{System Extension.} The current implementation of MUBox focuses on unlearning methods for image classification and deep learning. 
Instead of expanding evaluations across more models and datasets, we focus on an in-depth analysis of a select set of architectures and unlearning methods, given the urgent need for unified evaluation and systematic analysis of unlearning methods across different configurations and scenarios. We emphasize that the modular design and implementation of the MUBox framework enables easy adaptation and extension to specific data preparation, model architectures required for different tasks and data modalitites, as well as new development of unlearning methods and evaluation metrics. While our discussion focuses on evaluating ResNet18 on CIFAR10,  
additional model APIs for VGG, MobileNet, and ViT, as well as dataset APIs for SVHN, CIFAR100, and ImageNet are provided in MUBox.

\subsection{Unlearning Scenarios}
\label{sec:unlean_scenarios}

To explore the full potential of existing unlearning methods, we consider the following six practical scenarios.

(1) \textbf{Forgetting from one class}. A subset of training data from a specific class is randomly selected to be unlearned.

(2) \textbf{Forgetting from all classes}. A subset of training data from all classes is randomly selected to be unlearned. 

(3) \textbf{Class-wise forgetting}. All data from a whole class are selected to be unlearned.

(4) \textbf{Worst-case forgetting}. A subset of data with the lowest loss in the original model is selected to be unlearned.

(5) \textbf{Best-case forgetting}. A subset of data with the highest loss in the original model is selected to be unlearned.

(6) \textbf{Depoisoning}. A subset of data is poisoned and used to attack the model. After that, they are selected as forgetting data to be unlearned from the poisoned model.

\subsection{Evaluation Metrics}
The success of unlearning relies on three critical factors. A practical method must (1) maintain the performance of the model on both the retaining dataset $\mathcal{D}_r$ and the test dataset $\mathcal{D}_t$; (2) effectively erase the influence of $\mathcal{D}_f$ from the original model; and (3) offer efficiency improvements over retraining. An approach that does not meet these criteria should be deemed ineffective, as it may compromise model performance, fail to adequately eliminate the data intended for forgetting, or be less efficient than retraining. In order to comprehensively evaluate unlearning methods in deep learning, we consider three sets of evaluation metrics.

\textbf{\textit{Model Utility}}. We evaluate the model utility of the unlearned model using three metrics: (1)\textit{ Test Accuracy (\textbf{TA})}: accuracy on test dataset $\mathcal{D}_t$; (2) \textit{Retaining Accuracy (\textbf{RA})}: accuracy on retain dataset $\mathcal{D}_r$. The unlearned model should preserve the accuracy levels of the retrained model on $\mathcal{D}_t$ and $\mathcal{D}_r$ to maintain the model utility.

\textbf{\textit{Unlearning Efficacy}}. We evaluate unlearning efficacy using three sets of key metrics: (1) \textit{Forgetting Accuracy (\textbf{FA})} measures the accuracy change of the unlearned model on the forgetting dataset $\mathcal{D}_f$ compared with the gold-standard Retrain model. A smaller FA indicates less unlearning performance disparity with the Retrain. (2) \textbf{MIA-Efficacy} on $D_f$ measures if the forgetting data in the unlearned model can be identified as training data by the Membership Inference Attack (MIA)~\cite{shokri2017membership}. MIA determines whether a specific data point was part of the model's training process, while machine unlearning seeks to remove the knowledge of forgetting data from the model. The effectiveness of MIA has been validated in prior research~\cite{shokri2017membership,yeom2018privacy,song2019privacy} and is commonly used as an evaluation metric in machine unlearning literature~\cite{jia2023model,chen2023boundary, GradientProjectionUnlearning, fan2024salun, foster2024fast,heng2024selective,kurmanji2023towards,chundawat2023can}, particularly when theoretical guarantees are lacking or insufficient to quantify the residual knowledge of forgetting data. Since MIA involves deducing membership by calculating various metrics on the prediction vectors, we assess MIA predictors trained on five metrics: correctness~\cite{leino2020stolen}, confidence~\cite{shokri2017membership,yeom2018privacy,song2019privacy}, entropy~\cite{shokri2017membership}, modified entropy (m\_entropy)~\cite{song2021systematic}, and probability vector~\cite{shokri2017membership,song2021systematic}. 
In detail, we train these MIA predictors based on the correctness of the model's predictions, the confidence level associated with the correct class, the output entropy of the model, the modified output entropy with the probability of the ground-truth class flipped, and the output probability vectors, respectively. We use the true negative rate to evaluate the MIA efficacy, i.e., the ratio of forgetting data that are correctly classified by the MIA predictor as ``unseen" data to the size of the forgetting dataset. If the unlearning is effective, we expect a higher MIA value. (3) \textbf{$\ell_2$ distance}~\cite{thudi2022unrolling,tarun2023fast,wu2020deltagrad} measures the $\ell_2$ distance between the model parameters of the retrained model  $\theta_u^{*}$ and those of the unlearned model $\theta_u$.

\textbf{\textit{Cost Efficiency}}. we measure the Run-Time Efficiency (\textbf{RTE}) of unlearning methods and compare it with that of \textbf{Retrain}. We also assess the additional associated \textbf{storage overhead}.

\begin{table*}[t]
\centering
\caption{Evaluation of CIFAR10 under the forgetting from one class scenario, by forgetting 20\% samples in class 0. TA, RA, and FA represent Test Accuracy, Retaining Accuracy, and Forgetting Accuracy, respectively. Note that higher or lower values cannot directly demonstrate the effectiveness of unlearning; all metrics should be considered and interpreted together. Favorable results of FA minimize performance disparity with the retrained model.}

\label{tab:random-one-cls}
\resizebox{0.90\textwidth}{!}{%
\begin{tabular}{c|cc|cccccc|c}
\toprule
\rowcolor{gray!20}
\textbf{Method} & {\textbf{TA}} & {\textbf{RA}} & {\textbf{FA}} & {\textbf{correctness}} & {\textbf{confidence}} & {\textbf{entropy}} & {\textbf{m\_entropy}} & {\textbf{prob.}} & {\textbf{$\ell_2$}($\downarrow$)} \\
\midrule
Original & 93.24 & 100.00 & 100.00 & 0.00 & 0.60 & 3.10 & 100.00 & 100.00 & - \\
\hline
\rowcolor{blue!20}Retrain & 93.03  & 100.00 & 93.40  & 6.60 & 13.70 & 18.80 &100.00 & 96.30 & 0.0000 \\
Amnesiac~\cite{heng2024selective} & 69.00 & 71.07 & -25.60 & 32.20 & 32.60 & 52.90 & 100.00 & 86.40 & 0.6698 \\
\rowcolor{gray!5}PGU~\cite{GradientProjectionUnlearning}      & 91.48 & 99.51 & -1.60 & 8.20 & 17.70 & 27.20 & 100.00 & 16.30 & 0.0208 \\
Unrolling~\cite{thudi2022unrolling} & 92.45 & 100.00 & +6.00 & 0.60 & 2.50 & 6.50 & 100.00 & 100.00 & 0.0464 \\ 
\rowcolor{gray!5}SCRUB~\cite{kurmanji2023towards} & 92.85 & 100.00 & +6.60 & 0.20 & 1.70 & 4.40 & 100.00 & 100.00 & 0.0361  \\ 
$\ell_1$-Sparsity~\cite{jia2023model} & 92.66 & 100.00 & +6.60 & 0.10 & 2.10 & 5.20 & 100.00 & 100.00 & 0.0942  \\ 
\rowcolor{gray!5}First-order~\cite{warnecke2023machine} & 92.01 & 99.76 & +4.10 & 2.50 & 8.60 & 17.40 & 100.00 & 99.80 & 0.0208  \\ 
Second-order~\cite{warnecke2023machine} & 93.23 & 100.00 & +6.60 & 0.00 & 0.60 & 3.20 & 100.00 & 100.00 & 0.0208 \\ 
\rowcolor{gray!5}SSD~\cite{foster2024fast} & 82.37 & 91.33 & -93.40 & 100.00 & 100.00 & 98.70 & 100.00 & 89.00 & 0.0208 \\ 
\rowcolor{gray!5}Bad-T~\cite{chundawat2023can}  & 92.53  & 100.00 & +6.60 & 0.40   & 100.00 & 99.90 & 100.00 & 92.80 & 0.0224 \\ 
SalUn~\cite{fan2024salun}   & 93.13  & 100.00 & +6.60 & 0.00   & 99.30  & 100.00 & 100.00 & 99.30 & 0.0719 \\ 

\rowcolor{gray!5}L-CODEC~\cite{Mehta_2022_CVPR} & 77.93 & 83.61 & -70.90 & 13.20 & 9.90 & 2.00 & 100.00 & 99.60 & 0.0724 \\ 
Boundary-S~\cite{chen2023boundary}  & 82.82  & 91.53  & -70.90  & 77.50  & 88.50  & 74.00 & 0.00 & 65.30 & 0.0219 \\ 
\rowcolor{gray!5}Boundary-E~\cite{chen2023boundary}  & 82.69  & 91.49  & -70.80  & 77.40  & 88.10  & 74.00 & 0.00 & 65.70 & 0.0219 \\

FCS~\cite{FCS}	& 91.57 & 99.53	&+5.00	&1.60 &9.00 &15.50	&100.00 &99.80	& 0.0621	\\
\rowcolor{gray!5}MSG~\cite{MSG} & 83.40 &	87.20 &	 -9.40 & 16.00	&20.60	&31.70	&0.00	&91.90	& 0.0464	\\
CT~\cite{CT}&	65.60	&66.64	&-30.60 &37.20	&25.80	&53.90	&100.00	&83.00	&0.0461	\\
\rowcolor{gray!5}NIU~\cite{NIU} & 89.33& 97.06& +3.40 & 3.20	& 8.60	& 19.90	& 100.00	& 99.80	& 0.0663 \\

\midrule
Original  & 83.82 & 88.91 & 17.50 & 82.50 & 76.60 & 86.90 & 100.00 & 89.90 & - \\ 
\rowcolor{gray!5}SISA~\cite{bourtoule2021machine}  & 83.80 & 89.02 & 15.40 & 84.60 & 80.23 & 91.50 & 100.00 & 87.70 & 0.0000 \\ 
\midrule
Original & 81.20 & 100.00 & 100.00 & 0.00 & 8.00 & 12.00 & 100.00 & 32.00 & - \\
\rowcolor{gray!5}Fisher~\cite{golatkar2020eternal}  & 57.00 & 61.00 &-45.50 & 52.00  & 60.00 & 80.00 & 0.00 & 56.00 & 0.0002 \\ 
NTK~\cite{golatkar2020forgetting}   & 80.80 & 99.17 & +6.60 & 0.00   & 0.00 & 15.00 & 100.00 & 100.00 & 0.0001 \\ 
\bottomrule
\end{tabular}
}

\end{table*}

\section{Evaluations}
\label{sec:evaluation}

In this section, we evaluate 23 unlearning methods in deep learning currently implemented in MUBox. For the experiments, we use the same unlearning hyperparameter settings as specified in each original paper. Unless otherwise specified, class 0 is selected as the default in forgetting from one class and class-wise forgetting scenarios. All experiments are conducted on 2 NVIDIA Tesla V100 GPUs, each with 32 GB memory. The evaluation results for CIFAR-10 are presented in the main paper, while similar results for TinyImageNet will be provided as supplementary material upon acceptance of the paper.

\subsection{Forgetting from One Class}
\label{sec:random_one_cls}

We first evaluate unlearning methods under the forgetting from one class scenario. Specifically, we randomly select 1000 (20\%) samples from class 0 as the forgetting dataset. \textbf{Table~\ref{tab:random-one-cls}} presents the result.
  
\textbf{Model Utility.} After retraining without the forgetting dataset, the gold-standard Retrain model achieves a TA of 93.03\% and an RA of 100.00\%, representing a slight decrease of 0.19\% in TA compared to the original model. When comparing unlearning methods to the Retrain baseline, several methods demonstrate minimal impact on model utility, including PGU, Unrolling, SCRUB, $\ell_1$-sparsity, First-order, Second-order, Bad-T, SalUn, and FCS. Specifically, PGU achieves 91.48\%, Unrolling achieves 92.45\%, SCRUB achieves 92.85\%,  $\ell_1$-sparsity achieves 92.66\%, First-order achieves 92.01\%, Secone-order achieves 92.23\%, Bad-T achieves 92.53\%, SalUn achieves 93.13\%, and FCS achieves 91.57\%. These methods also maintain RA with little to no degradation, indicating well-preserved model utility after unlearning. SISA also maintains model utility with a TA of 83.8\% and an RA of 89.02\%, which is expected since it retrains the shard model from scratch.

In contrast, some unlearning approaches compromise the model performance on the forgetting class, even when only 20\% of the data in the class is unlearned. For instance, Amnesiac reduces the TA and RA to 69.00\% and 71.01\%, respectively. SSD drops to 82.37\% TA and 91.33\% RA, and L-CODEC has 77.93\% TA and 83.61\% RA. MSG decreases the TA and RA to 83.4\% and 87.2\%, and CT has only TA and RA of 65.6\% and 66.64\%, respectively. Boundary-E and Boundary-S show similar results, with all these methods demonstrating at least a 10\% accuracy drop compared to the Retrain baseline. 

This performance loss can be attributed to the design of these algorithms. For example, Amnesiac requires saving and later adding back parameter updates from batches containing data that need potential removal, which can lead to significant degradation in model performance when repeatedly encountering forgetting data in training batches. 

For Fisher and NTK, we follow their experimental setting given their theoretical assumptions and the need for additional datasets~\cite{golatkar2020forgetting,golatkar2020eternal}. Compared with the performance of their original model, NTK maintains model utility with 99.17\% RA and 80.80\% TA, while Fisher significantly degrades performance, dropping TA and RA to 57.00\% and 61.00\%, respectively. Fisher's poor performance may be due to its sensitivity to hyperparameter settings. Extensive tuning is required to adapt to different models, datasets, and unlearning scenarios, thereby limiting its practical applicability.

\begin{remark}
\textbf{Remark 2.} \textit{Existing unlearning methods, when applied outside their original experimental environment, often experience significant performance loss, highlighting their lack of robustness across different settings.}
\end{remark}

\textbf{Unlearning Efficacy.}
Table~\ref{tab:random-one-cls} also presents FA and MIA scores on the forgetting dataset. Regarding FA, unlearning methods that maintain high TA and RA generally also show higher FA compared to Retrain. Methods such as Unrolling (+6.00\%), SCRUB (+6.60\%), $\ell_1$-sparsity (+6.60\%), First-order (+4.1\%), Second-order (+6.60\%), Bad-T (+6.60\%), SalUn (+6.60\%), FCS (+5.00\%), NIU (+3.40\%) remain such trend. In contrast, methods that compromise model utility exhibit a significant reduction in FA. For instance, Amnesiac (-25.6\%) and CT (-30.6\%) show notable decreases. More extreme cases include SSD (-93.40\%), Boundary-S (-70.90\%), Boundary-E (-70.80\%), and L-CODEC (-70.90\%), which substantially lower the accuracy on the forgetting data. Notably, PGU demonstrates the lowest discrepancy compared to Retrain, with only a 1.6\% decrease in FA. It is important to note that FA alone may not accurately capture the effectiveness of unlearning when only part of a class's data is forgotten from one class, as data from the same class in the retaining data may be generalized to the forgetting data.

Regarding MIA scores, a number close to 100.00\% indicates that all forgetting data are classified as ``unseen", and vice versa. Before unlearning, the original MIA scores should be low. We make two observations on the MIA score-based measurement. First, MIA scores derived from m\_entropy and probability are unreliable in the forgetting from one class scenario, and they should not be used to evaluate unlearning performance in this set of experiments. 
Second, MIA scores vary significantly across different unlearning methods. Before unlearning, the MIA scores of the original model in terms of correctness, confidence, and entropy are 0.00\%, 0.60\%, and 3.10\%, respectively. After retraining without the forgetting data, the MIA scores for correctness, confidence, and entropy increase by 6.60\%, 13.70\%, and 18.80\%, respectively. Even though the forgetting data is removed from the retraining, correctness, confidence, and entropy cannot reliably reflect the unlearning efficacy of the Retrain approach. PGU, First-order, and FCS can achieve MIA scores close to those of Retrain. Most unlearning approaches we evaluate demonstrate inconsistent unlearning efficacy across different MIA scores compared to Retrain.
For example, Unrolling, SCRUB,  $\ell_1$-sparsity, First-order, and Second-order has high unlearning efficacy under probability-based MIA measurement but not on correctness, confidence, and entropy. Their MIA scores are also close to those of the original models. Bad-T and SalUn reach near 100\% in terms of confidence, entropy, and probability but fail badly to 0.40\% and 0.00\% under the correctness metric. Only SSD and SISA consistently deliver high unlearning efficacy across all five metrics. However, as discussed earlier, these high MIA scores come at the cost of reduced model utility. 
This demonstrates the limitation of previous machine unlearning works~\cite{jia2023model, GradientProjectionUnlearning, graves2021amnesiac} that evaluate MIA based on only one of these aspects, potentially leading to biased evaluations. 

\begin{remark}
\textbf{Remark 3.} \textit{
While FA and MIA are key metrics for evaluating unlearning efficacy, relying on a single metric can result in a biased assessment.}
\end{remark}

In addition to MIA scores, the last column of Table~\ref{tab:random-one-cls} presents the normalized $\ell_2$ distance between the retrained model and the unlearned model obtained from different unlearning methods. Intuitively, the $\ell_2$ distance measures how similar the unlearned model is to the retrained model. A smaller $\ell_2$ distance suggests that the models should behave similarly after unlearning~\cite{tarun2023fast,wu2020deltagrad,thudi2022unrolling}. However, Table~\ref{tab:random-one-cls} shows that $\ell_2$ distance cannot reflect actual differences in model behavior or performance after unlearning. For instance, SSD and PGU have the same $\ell_2$ distances (0.0208 and 0.0208, respectively), which would imply similar model behavior. Nonetheless, SSD severely compromises model utility, while PGU preserves it. In addition, using $\ell_2$ distance for unlearning efficacy evaluation overlooks the complexity of high-dimensional parameter spaces and the non-linearity of deep learning models, where different local minima can yield models with comparable behavior despite differences in parameter values. 

\begin{remark}
\textbf{Remark 4.} \textit{
Although MIA scores may provide a more aligned measure of unlearning efficacy, there is no definitive gold standard for quantitatively assessing these methods. Designing an effective metric remains non-trivial.}
\end{remark}

\begin{table}[t]
\centering
\caption{Measurement of time cost compared to the Retrain baseline and additional storage requirement.}
\label{tab:cost_efficiency}
\resizebox{0.75\columnwidth}{!}{%
\begin{tabular}{c|cc}
\toprule
\rowcolor{gray!20}\textbf{Method} & \textbf{RTE}($\downarrow$) & \textbf{Storage}(GB)($\downarrow$) \\ 
\midrule
Amnesiac & 0.39$\times$ & 427.99  \\
PGU      & 0.16$\times$ & 0.72\\
Unrolling & 0.04$\times$ & 0.00\\ 
SCRUB &  0.03$\times$ & 0.00\\ 
$\ell_1$-Sparsity & 0.06$\times$ & 0.00  \\ 
First-order & 0.00008$\times$ & 0.00 \\ 
Second-order & 0.012$\times$ & 0.00 \\ 
SSD & 0.008$\times$ & 0.00 \\ 
Bad-T  & 0.004$\times$ & 0.00 \\ 
SalUn   & 0.08$\times$ & 0.09 \\ 
SISA   & 0.965$\times$ &  0.22 \\ 
L-CODEC & 8.198$\times$ & 0.00 \\ 
Boundary-S & 0.002$\times$ & 0.00 \\ 
Boundary-E & 0.002$\times$ & 0.00 \\
FCS & 0.038$\times$ & 0.00\\
MSG & 0.024$\times$& 0.00\\
CT & 0.022$\times$& 0.00\\
NIU & 0.04$\times$& 0.00 \\
Fisher  & 0.016$\times$ & 0.00 \\ 
NTK   & 0.046$\times$ & 35.91 \\ 
\hline
UNSIR & 0.008$\times$ & 0.00 \\
GKT & 0.199$\times$ & 0.88 \\
\bottomrule
\end{tabular}
}
\end{table}

\begin{table*}[t]
\centering
\caption{Evaluation on CIFAR10 under the forgetting from all classes scenario.}
\label{tab:random-all-cls}
\resizebox{0.9\textwidth}{!}{%
\begin{tabular}{c|cc|cccccc|c}
\toprule
\rowcolor{gray!20}
\textbf{Method} & {\textbf{TA}} & {\textbf{RA}} & {\textbf{FA}} & {\textbf{correctness}} & {\textbf{confidence}} & {\textbf{entropy}} & {\textbf{m\_entropy}} & {\textbf{prob.}} & {\textbf{$\ell_2$}($\downarrow$)} \\
\midrule
Original  & 93.24 & 100.00 & 100.00 & 0.00 & 0.40 & 1.90 & 11.90 & 0.10 & - \\
\hline
\rowcolor{blue!20}Retrain & 92.93  & 100.00 & 90.10 & 9.90 & 19.00 & 23.50 & 31.40 & 26.60 & 0.0000 \\
Amnesiac~\cite{heng2024selective} & 69.70 & 71.51 & -20.78 & 55.20 & 64.30 & 78.00 & 54.70 & 43.10 & 0.6698  \\
\rowcolor{gray!5}PGU~\cite{GradientProjectionUnlearning}      & 92.81 & 100.00 & +4.60 & 5.30 & 6.00 & 7.10 & 16.10 & 1.50 & 0.0209 \\
Unrolling~\cite{thudi2022unrolling} & 93.17 & 100.00 & +9.90 & 0.00 & 0.30 & 2.20 & 17.70 & 10.60 & 0.0464 \\
\rowcolor{gray!5}SCRUB~\cite{kurmanji2023towards} & 93.26 & 100.00 & +9.90 & 0.00 & 0.50 & 2.00 & 14.00 & 10.60 & 0.0360 \\
$\ell_1$-Sparsity~\cite{jia2023model} & 92.75 & 100.00 & +9.90 & 0.10 & 1.70 & 4.00 & 14.80 & 0.90 & 0.0942 \\
\rowcolor{gray!5}First-order~\cite{warnecke2023machine} & 92.90 & 100.00 & +9.70 & 0.20 & 0.80 & 2.20 & 13.70 & 21.90 & 0.0208  \\
Second-order~\cite{warnecke2023machine} & 93.24 & 100.00 & +9.90 & 0.00 & 0.40 & 2.00 & 11.90 & 0.10 & 0.0208 \\ 
\rowcolor{gray!5}SSD~\cite{foster2024fast} & 50.14 & 52.72 & -38.10 & 48.00 & 44.90 & 19.60 & 46.10 & 58.90 & 0.0208 \\
\rowcolor{gray!5}Bad-T~\cite{chundawat2023can}  & 92.42 & 100.00 & +8.60 & 1.30 & 11.20 & 22.70 & 30.10 & 10.90 &0.0224  \\
SalUn~\cite{fan2024salun}   & 92.64 & 100.00 & +8.50 & 1.40 & 15.00 & 17.70 & 40.80 & 7.70 & 0.0719 \\

\rowcolor{gray!5}L-CODEC~\cite{Mehta_2022_CVPR} & 69.55 & 75.00 & -54.00 & 63.90 & 61.70 & 57.00 & 57.70 & 84.00 & 0.0724 \\
Boundary-S~\cite{chen2023boundary}  & 93.17 & 100.00 & +9.90 & 0.00 & 0.50 & 2.20  & 10.70 & 0.10 & 0.0219 \\
\rowcolor{gray!5}Boundary-E~\cite{chen2023boundary}  & 93.20 & 100.00 & +9.90 & 0.00 & 0.50 & 2.10 & 16.10 & 0.10 & 0.0219 \\
FCS~\cite{FCS}	&92.28	&99.87	&+9.80 &0.10	&5.10	&06.30	&11.90	&03.60	&0.0713	\\	
\rowcolor{gray!5}MSG~\cite{MSG}	&84.20	&88.00	&-2.20 &12.10	&13.90	&21.10	&43.80	&49.90	&0.0464	\\	
CT~\cite{CT}	&65.55	&66.56	&-21.90 &31.80	&26.30	&44.30	&45.80	&33.90	&0.0413	\\	
\rowcolor{gray!5}NIU~\cite{NIU} &89.52	&97.10	&+7.50 &2.40	&7.50	&15.80	&24.00	&13.20	&0.0576	\\	
\midrule
Original  & 83.32 & 88.84 & 89.40 & 10.60 & 4.50 & 26.50 & 48.80 & 49.10 & - \\ 
\rowcolor{gray!5}SISA~\cite{bourtoule2021machine}  & 84.02 & 88.99 & 85.40 & 14.60 & 9.00 & 47.30 & 42.10 & 50.00 & 0.0000 \\ 
\midrule
Original & 81.20 & 100.00 & 100.00 & 0.00 & 4.00 & 12.00 & 8.00 & 4.00 & - \\
\rowcolor{gray!5}Fisher~\cite{golatkar2020eternal}  & 55.80 & 61.50 & -42.10 & 52.00 & 52.00 & 60.00 & 40.00 & 44.00 & 0.0002 \\
NTK~\cite{golatkar2020forgetting}   & 81.40 & 99.18 & +9.90 & 0.0 & 0.00 & 0.00 & 20.00 & 0.00 & 0.0001 \\
\bottomrule
\end{tabular}
}
\end{table*}

\textbf{Cost Efficiency.} \textbf{Table~\ref{tab:cost_efficiency}} presents the RTE compared with Retrain and the additional storage overhead required by each unlearning method. Regarding RTE, almost all approximate unlearning evaluated, except L-CODEC, offers a significant reduction compared with Retrain. L-CODEC incurs substantial cost of time due to its need to identify the subset of model parameters to be updated for each forgetting data. Moreover, while SISA is tailored to reduce the retraining cost for exact unlearning, it achieves only a 0.965$\times$ runtime efficiency, almost equivalent to Retrain. This limited efficiency may be due to the trade-off between unlearning budget and retraining efficiency. As stated in~\cite{bourtoule2021machine}, SISA can exhibit speed-up only when unlearning requests $K < 3S$, where $S$ denotes the number of shards. However, using more shards degrades the test accuracy. In our experiments, even with only five shards, the TA of SISA before unlearning is merely 83.82\%, nearly 10\% lower than that of the original model (93.24\%).

Regarding storage costs, methods like Amnesiac, PGU, SISA, SalUn, and NTK require additional storage overhead. Specifically, Amnesiac requires 427.99 GB to store the information used for unlearning because it needs to save almost all parameter updates at each training step. SISA requires 0.22GB to save checkpoints of shard models (and more if the shard datasets are further sliced), while SalUn needs an extra 0.09GB storage to store the salient weights for unlearning. NTK faces a computational bottleneck as the size of the NTK matrix grows exponentially with the number of training samples and classes, making in-memory computations impractical and requiring 35.91 GB of disk space to store the essential information needed for computation.

\subsection{Forgetting from All Classes}

\label{sec:random_all_cls}

We next consider the scenario of forgetting from all classes. Specifically, we randomly select 1000 samples from all classes as the forgetting dataset. \textbf{Table~\ref{tab:random-all-cls}} presents the results.

\textbf{Model Utility}. 
Similar to the forgetting from one class scenario earlier, most unlearning methods in our evaluation, including PGU, Unrolling, SCRUB, $\ell_1$-Sparsity, First-order, Second-order, Bad-T, SalUn, NTK, Boundary-S, Boundary-E, FCS, and SISA maintain model utility well in terms of TA and RA. It is worth noting that Boundary-S and Boundary-E, which underperformed when forgetting from one class, only slightly decreased the TA by 0.07\% and 0.04\%, respectively, compared with the original model, while securing 100\% RA. These methods effectively preserve model utility.

Still, Amnesiac severely comprises model utility, with TA and RA dropping to 69.70\% and \%71.51, respectively. 
L-CODEC also underperforms, achieving 69.55\% TA and 75.00\% RA due to its lack of robustness against varying unlearning budgets. Similarly, SSD and Fisher degrade model utility, with SSD achieving 50.14\% TA and 52.72\% RA, which is worse than the forgetting from one class scenario. Fisher achieves 55.80\% TA and 61.50\% RA. Both SSD and Fisher rely heavily on the Fisher Information Matrix and are sensitive to hyperparameters.

\begin{table*}[t]
\centering
\caption{Evaluation on CIFAR10 under the class-wise forgetting scenario. GTK and UNSIR are class-wise forgetting only.}
\label{tab:class}
\resizebox{0.9\textwidth}{!}{%
\begin{tabular}{c|cc|cccccc|c}
\toprule
\rowcolor{gray!20}
\textbf{Method} & {\textbf{TA}} & {\textbf{RA}} & {\textbf{FA}} & {\textbf{correctness}} & {\textbf{confidence}} & {\textbf{entropy}} & {\textbf{m\_entropy}} & {\textbf{prob.}} & {\textbf{$\ell_2$}($\downarrow$)} \\
\midrule
Original & 93.24 & 100.00 & 100.00 & 0.02 & 0.30 & 2.50 & 100.00 & 100.00 & - \\
\hline
\rowcolor{blue!20}Retrain & 84.15  & 100.00 & 0.00 & 100.00 & 100.00 & 62.38 & 100.00 & 90.28 & 0.0000  \\
Amnesiac~\cite{heng2024selective}  & 57.67 & 65.09 & 0.00 & 100.00 & 100.00 & 76.64 & 100.00 & 86.66  & 0.6716 \\
\rowcolor{gray!5}PGU~\cite{GradientProjectionUnlearning}       & 90.67 & 100.00 & +85.14 & 8.38 & 43.06 & 69.30 & 100.00 & 95.50 & 0.1138 \\		 				
Unrolling~\cite{thudi2022unrolling} & 83.41 & 100.00 & 0.00 & 100.00 & 100.00 & 41.70 & 0.00 & 72.82 & 0.1415 \\ 	 					
\rowcolor{gray!5}SCRUB~\cite{kurmanji2023towards} &	83.58 & 100.00	& 0.00 & 100.00 & 100.00 &	57.38 &	100.00 & 65.50 & 0.1298  \\ 
$\ell_1$-Sparsity~\cite{jia2023model} & 91.78 & 100.00 & +97.00 & 3.00 & 23.10 & 36.40 & 100.00 & 100.00 & 0.0924  \\ 
\rowcolor{gray!5}First-order~\cite{warnecke2023machine} & 82.68	& 99.40	& +5.60 & 94.36 & 99.40	& 67.06	& 100.00 & 82.08 & 0.1136  \\ 
Second-order~\cite{warnecke2023machine} & 93.21 & 100.00 & +100.00 & 0.02 & 0.30 & 2.90 & 100.00 & 100.00 & 0.1136 \\ 
\rowcolor{gray!5}SSD~\cite{foster2024fast} & 83.81 & 100.00 & 0.00 & 100.00 & 100.00 & 99.28 & 100.00 & 90.28 & 0.1136 \\ 
\rowcolor{gray!5}Bad-T~\cite{chundawat2023can}  & 83.68  & 100.00 & +1.50 & 98.50   & 100.00 & 100.00 & 100.00 & 100.00 & 0.1157 \\ 
SalUn~\cite{fan2024salun}  & 84.33 & 100.00 & +3.48  & 96.52 & 100.00 & 100.00 & 100.00 & 100.00 & 0.1693 \\
\rowcolor{gray!5}L-CODEC~\cite{Mehta_2022_CVPR} & 10.00 & 11.13  & 0.00  & 100.00& 100.00 & 100.00 & 0.00   & 100.00 & 0.1624 \\
Boundary-S~\cite{chen2023boundary} & 83.05 & 98.20  & +21.76 & 78.24 & 92.38  & 84.24  & 100.00 & 65.36  & 0.1193 \\
\rowcolor{gray!5}Boundary-E~\cite{chen2023boundary} & 83.36 & 98.40  & +23.68 & 76.32 & 92.04  & 85.04  & 100.00 & 67.66  & 0.1193 \\
FCS~\cite{FCS}	&83.42	&100.00& 0.00 & 100.00	& 100.00 & 100.00 & 98.98	& 96.48 &	0.1633	\\	
\rowcolor{gray!5}MSG~\cite{MSG}	&77.31	& 87.90	& +10.50 & 89.50	& 93.16	& 85.12	& 100.00	& 92.94 & 0.1414	\\	
CT~\cite{CT}	&60.92	& 68.34	& +3.42 & 96.58	& 87.65	& 53.92	& 72.27	& 78.32 & 0.1368	\\	
\rowcolor{gray!5}NIU~\cite{NIU} & 81.05	& 97.16	& 0.00	& 98.64 & 96.71	& 89.57& 86.97& 89.74 &	0.1586	\\	
GKT~\cite{chundawat2023zero}  & 9.80  & 11.08  & 0.00  & 100.00& 100.00 & 53.94  & 98.16  & 64.68  & 0.9997 \\
\rowcolor{gray!5}UNSIR~\cite{tarun2023fast} & 89.21 & 100.00 & +70.46 & 29.54 & 88.42  & 96.02  & 100.00 & 94.72  & 0.1218 \\
\midrule
Original  & 83.82 & 87.51 & 17.12 & 82.88 & 73.04 & 63.42 & 0.00 & 35.70 & - \\
\rowcolor{gray!5}SISA~\cite{bourtoule2021machine}   & 76.06 & 89.45  & 0.00  & 100.00& 100.00 & 88.94  & 100.00 & 80.82  & 0.0000 \\
\midrule
Original  & 83.82 & 100.00 & 100.00 & 0.00 & 7.00 & 10.00  & 0.00 & 100.00 & - \\
\rowcolor{gray!5}Fisher~\cite{golatkar2020eternal} & 58.20 & 65.30 & +51.00 & 51.00 & 73.00  & 79.00 & 6.00 & 70.00 & 0.0002 \\ 
NTK~\cite{golatkar2020forgetting}  & 79.4 & 99.25 & +90.00 & 15.00 & 49.00  & 58.00 & 100.00 & 99.00 & 0.0001 \\ 
\bottomrule
\end{tabular}
}
\end{table*}

\textbf{Unlearning Efficacy}. The FA results for most methods are consistent with those in Table~\ref{tab:random-one-cls}, where positive discrepancies in FA are observed for many methods, and negative discrepancies are typically associated with a partial or complete compromise in model utility. However, it is worth noting that while Boundary-S and Boundary-E maintain model utility under this scenario, they still exhibit +9.90\% in FA compared to Retrain. Similar results can be observed for FCS and NIU. This highlights that FA alone cannot effectively measure unlearning efficacy, as it primarily reflects model performance when partially forgetting data from classes. Similarly, the normalized $\ell_2$ distance between the retrained and unlearned models again proves to be an impractical metric for evaluating unlearning efficacy.

Regarding MIA scores, unlike in the forgetting from one class scenario, the MIA scores are more indicative, as shown in Table~\ref{tab:random-all-cls}. 
Before unlearning, the MIA scores of the original model are low across all 5 metrics. In contrast, after unlearning with Retrain,  the MIA scores increase to 9.90\%, 19.00\%, 23.50\%, 31.40\%, and 26.60\%, respectively, suggesting that while unlearning with Retrain would increase MIA scores, it still falls short of reliably classifying the forgetting data as unseen. Due to the same reason, although most of the unlearning methods in our evaluation preserve the model utility, their MIA scores remain close to those of the original model and Retrain, and are consistently low. In addition, the relatively higher MIA scores in Amnesiac, SSD, CT, SISA, L-CODEC, and Fisher are at the cost of compromising the model utility. 

Compared to the forgetting from one class scenario, both TA and RA are higher when forgetting from all classes, indicating better model utility when forgetting from all classes. However, most unlearning methods tend to have lower MIA scores, indicating generally poorer unlearning efficacy.

\begin{remark}
\textbf{Remark 5.} \textit{Forgetting from all classes proves significantly more challenging than forgetting from a single class, revealing that existing unlearning techniques are considerably less effective in the multi-class scenario. }
\end{remark}

\textbf{Cost Efficiency}. The results are similar to Table~\ref{tab:cost_efficiency} given the same 1000 total unlearning data.

\subsection{Class-wise Forgetting}

Next, we evaluate the performance of unlearning methods under the class-wise forgetting scenario, which is the most popular scenario in existing unlearning studies. Specifically, we select all images from class 0 (5000 images in total) as the forgetting dataset. Two additional unlearning approaches, GKT and UNSIR, are added as they are by-design class-wise unlearning methods. The results are presented in \textbf{Table~\ref{tab:class}}. 

\textbf{Model Utility.} Our results demonstrate that most unlearning methods are highly effective at preserving model utility while successfully unlearning the target class in the class-wise forgetting. Since we unlearn the entire class, the TA of Retrain drops to 84.15\%, a 9.09\% decrease from the original TA as the class is wiped out. The TA of Unrolling, SCRUB, First-order, SSD, Bad-T, SalUn, Boundary-S, Boundry-E, FCS, and NIU are all close to that of Retrain. 
Additionally, these methods maintain an RA close to 100.00\%, with only Boundary-E and Boundary-S showing a slight decrease in RA.

\textbf{Unlearning Efficacy.} Regarding FA, methods including Unrolling, SCRUB, First-order SSD, Bad-T, SalUn, FCS, NIU and SISA exhibit minimal FA discrepancies, remaining close to zero while maintaining competitive while maintaining competitive TA and RA results, indicating the (almost) complete removal of the forgetting class. By comparison, Boundary-S and Boundary-E exhibit relatively higher FA discrepancies of +21.76\% and +23.68\%, respectively, compared to Retrain. While methods like PGU, $\ell_1$-Sparsity, Second-order, NTK, and UNSIR have high TA and RA, their FA results imply less successful unlearning efficacy, which is further confirmed in their MIA scores. For methods like Amnesiac (0.00\%), L-CODEC (0.00\%), GKT (+0.00\%), and CT (+3.42\%), their low FA discrepancies are not necessarily indicative of effective unlearning. Instead, they are the results of model failure, as evidenced by significantly degraded TA and RA.

\begin{figure}[!t]
\centering
\includegraphics[width=\columnwidth]{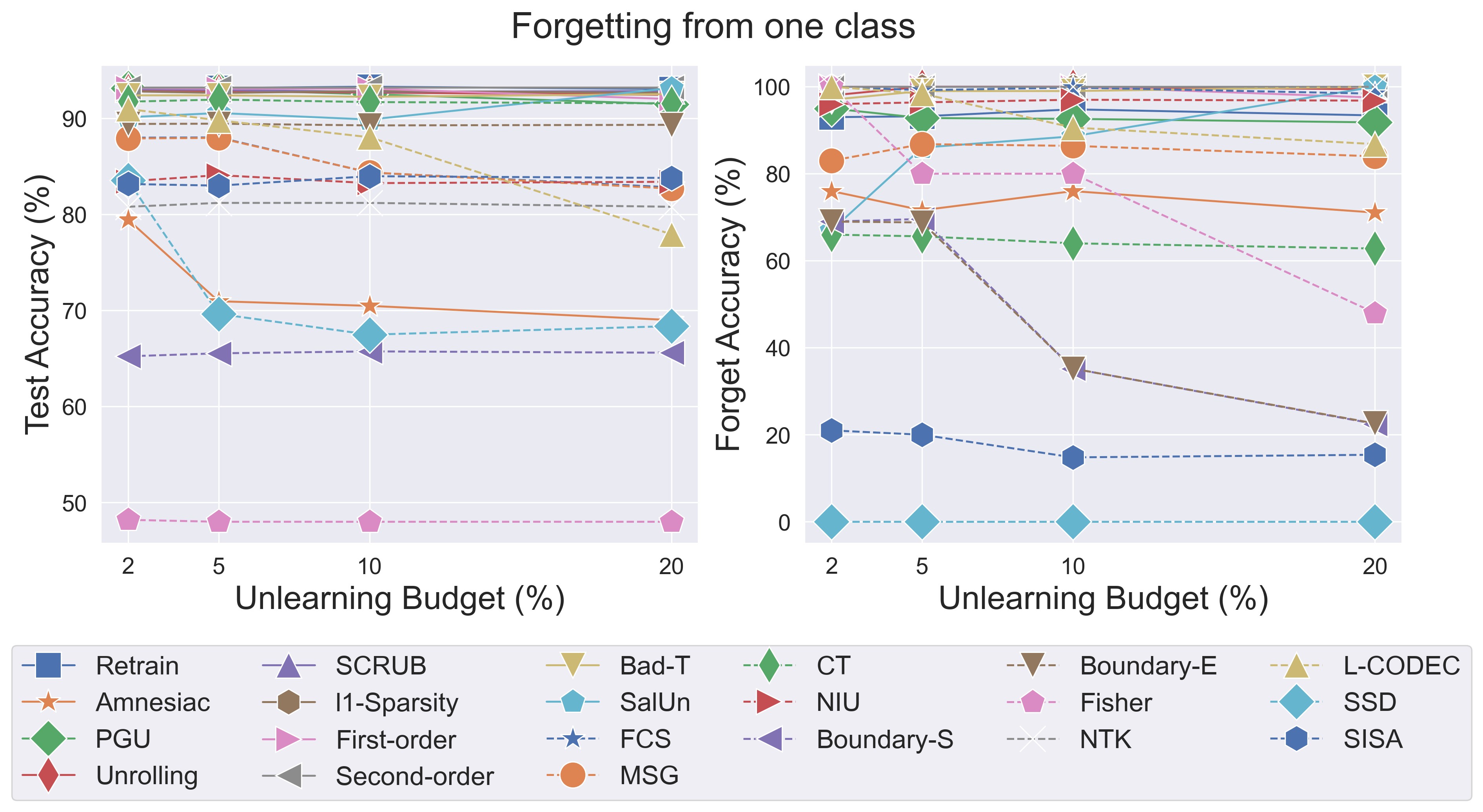}
\caption{Unlearning methods across different unlearning budgets in terms of test accuracy and forgetting accuracy  under the forgetting from one class scenario.}
\label{fig:budget_one_cls}
\end{figure}

In terms of MIA scores, since all data from class 0 are removed, the correctness, confidence, entropy, and probability-based MIA scores of Retrain are 100.00\%, 100.00\%, 62.38\%, and 90.28\%, respectively. M\_entropy is not used due to its failure as an indicator for unlearning efficacy. For methods that preserve the model utility, including Unrolling, SCRUB, First-order, SSD, Bad-T, SalUn, FCS, NIU, and SISA, they successfully perform the class-wise unlearning, with the latter two being slightly less effective, echoing the FA results. In contrast, methods like PGU, $\ell_1$-Sparsity, Second-order, NTK, and UNSIR show much worse MIA scores, implying their failure in unlearning. However, for methods that compromise the model utility entirely (Amnesiac, L-CODEC,  Fisher, and GTK), their MIA scores are notably high, even closing to 100\% in at least one of the MIA metrics. These observations confirm the challenge of quantitatively assessing unlearning efficacy.

\textbf{Cost Efficiency.} For RTE, the results in this scenario are similar to those in Table~\ref{tab:cost_efficiency}, except for L-CODEC, which shows a 5$\times$ increase in runtime due to its need to identify the subset of model parameters for each forgetting data point. While other methods show a slight increase in runtime, they still significantly improve over Retrain. For cost storage, results are also similar to Table~\ref{tab:cost_efficiency}. The additional storage required for PGU, SalUn, SISA, and NTK is mainly due to the need for model checkpoints or computations involving the entire dataset. Amnesiac continues to consume similar storage space. With a large unlearning budget, it saves nearly all parameter updates at each training step.

\begin{remark}
\textbf{Remark 6.} \textit{Most unlearning methods are effective in maintaining model utility and unlearning in class-wise forgetting, which has been the primary focus of the literature. This emphasis
potentially overlooks more challenging and complex unlearning scenarios. 
}
\end{remark}

\begin{figure}[!t]
\centering
\includegraphics[width=\columnwidth]{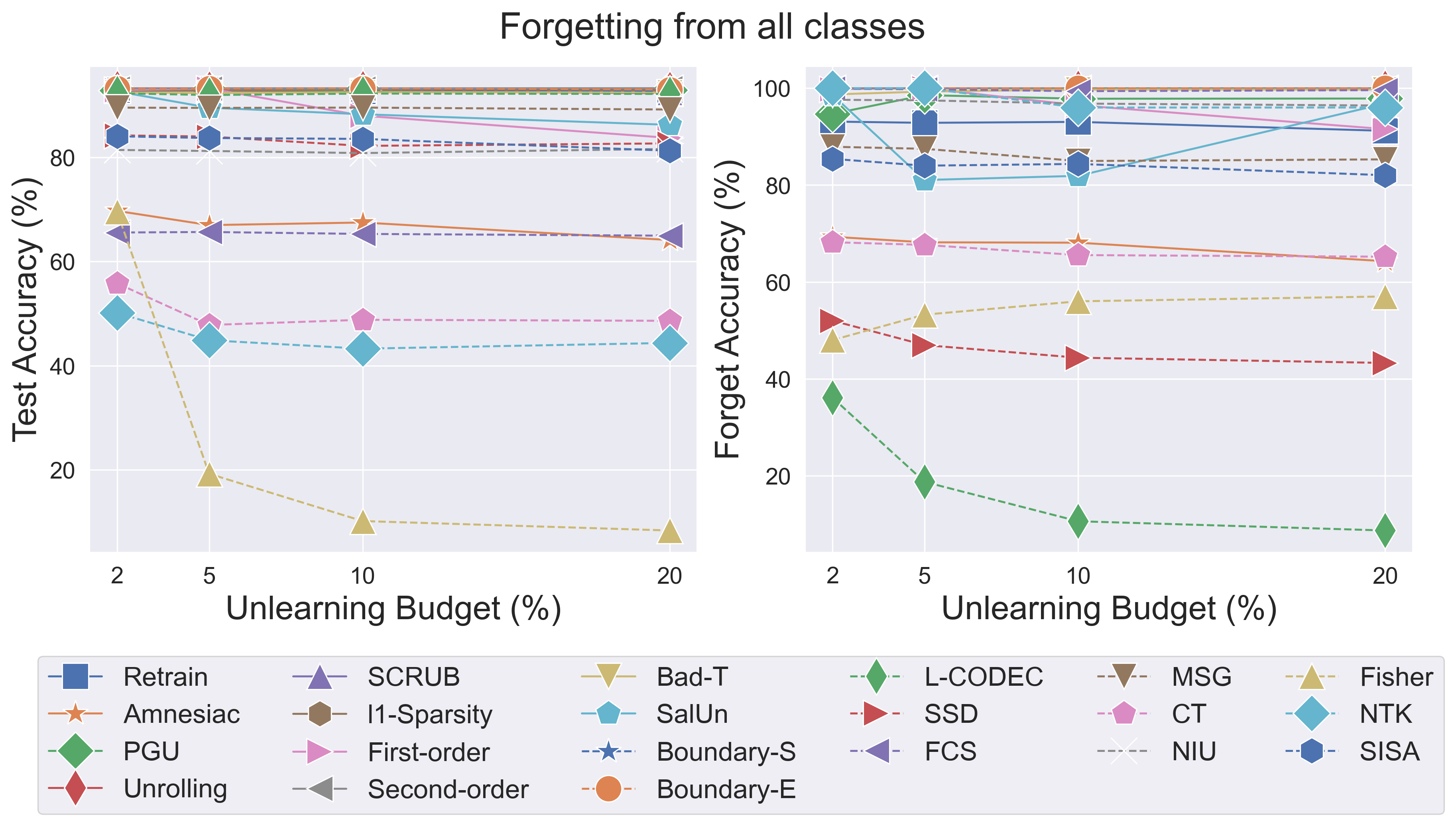}
\caption{Unlearning methods across different unlearning budgets in terms of test accuracy and forgetting accuracy under the forgetting from all classes scenario.}
\label{fig:budget_all_cls}
\end{figure}

\subsection{Impact of Unlearning Budget}

We further evaluate the performance of various unlearning methods across different unlearning budgets under both the forgetting from one class and forgetting from all classes scenarios. Specifically, we experiment with unlearning budgets of 2\%, 5\%, 10\%, and 20\% for both scenarios, i.e., 100, 250, 500, 1000 samples for forgetting from one class and 1000, 2500, 5000, and 10000 samples for forgetting from all classes. Their results are illustrated in \textbf{Figure~\ref{fig:budget_one_cls}} and \textbf{Figure~\ref{fig:budget_all_cls}}, respectively.

\begin{table*}[htbp]
\centering
\caption{Evaluation on CIFAR10 under the worst-case forgetting scenario.}
\label{tab:worst-case}
\resizebox{0.88\textwidth}{!}{%
\begin{tabular}{c|cc|cccccc|c}
\toprule
\rowcolor{gray!20}
\textbf{Method} & {\textbf{TA}} & {\textbf{RA}} & {\textbf{FA}} & {\textbf{correctness}} & {\textbf{confidence}} & {\textbf{entropy}} & {\textbf{m\_entropy}} & {\textbf{prob.}} & {\textbf{$\ell_2$}($\downarrow$)} \\
\midrule
Original & 93.24 & 100.00 & 100.00 & 0.00 & 0.00 & 0.00 & 74.80 & 17.20 & - \\
\hline
\rowcolor{blue!20}Retrain  & 92.82  & 100.00 & 93.10 & 6.90 & 12.10 & 15.50 & 26.70 & 16.30 & 0.0000 \\
Amnesiac~\cite{heng2024selective} & 68.86 & 71.69 & -52.70 & 59.60 & 78.90 & 87.90 & 58.70 & 66.70 & 0.6698   \\
\rowcolor{gray!5}PGU~\cite{GradientProjectionUnlearning}      & 92.51 & 100.00 & +6.90 & 0.10 & 0.20 & 0.20 & 28.00 & 17.30 & 0.0209 \\
Unrolling~\cite{thudi2022unrolling} & 92.49 & 100.00 & +6.90 & 0.00 & 0.20 & 0.90 & 15.90 & 17.20 & 0.0464 \\
\rowcolor{gray!5}SCRUB~\cite{kurmanji2023towards} & 93.01 & 100.00 & +6.90 & 0.00 & 0.01 & 0.02 & 18.10 & 17.20 & 0.0361 \\
$\ell_1$-Sparsity~\cite{jia2023model} & 92.85 & 100.00 & +6.90 & 0.00 & 0.00 & 0.00 & 3.40 & 0.00 & 0.0942 \\
\rowcolor{gray!5}First-order~\cite{warnecke2023machine} & 91.77 & 99.50 & +6.90 & 0.10 & 0.80 & 1.30 & 41.40 & 1.40 & 0.0208 \\
Second-order~\cite{warnecke2023machine} & 93.24 & 100.00 & +6.90 & 0.00 & 0.00 & 0.00 & 75.30 & 17.20 & 0.0208  \\ 
\rowcolor{gray!5}SSD~\cite{foster2024fast} & 82.01 & 88.76 & -26.90 & 33.80 & 63.90 & 75.50 & 83.20 & 45.20 & 0.0208 \\
\rowcolor{gray!5}Bad-T~\cite{chundawat2023can}  & 92.32 & 99.80 & +6.30 & 0.60 & 71.80 & 80.00 & 83.60 & 62.20 & 0.0224 \\
SalUn~\cite{fan2024salun}   & 92.84 & 100.00 & +6.90 & 0.30 & 53.20 & 75.60 & 85.70 & 43.50 & 0.0719 \\
\rowcolor{gray!5}L-CODEC~\cite{Mehta_2022_CVPR} & 68.36 & 73.11 & +1.80 & 5.20 & 4.20 & 16.40 & 50.30 & 60.70 & 0.0724 \\
Boundary-S~\cite{chen2023boundary}  & 86.13 & 94.34 & -8.60 & 15.50 & 28.50 & 43.90 & 71.90 & 22.50 & 0.0219 \\
\rowcolor{gray!5}Boundary-E~\cite{chen2023boundary}  & 86.13 & 94.39 & -7.80 & 14.70 & 27.60 & 43.00 & 70.20 & 22.10 & 0.0219 \\
FCS~\cite{FCS}	&92.12	&99.85	&+6.80 &0.10	&0.60	&0.80	&7.30	&0.40	&0.0626	\\
\rowcolor{gray!5}MSG~\cite{MSG}	&82.79	&87.00	&-15.00 &21.90	&30.20	&62.90	&56.70	&81.70	&0.0464	\\	
CT~\cite{CT}	&65.11	&66.81	&-34.30 &41.20	&51.90	&50.40	&42.20	&63.90	&0.0458	\\	
\rowcolor{gray!5}NIU~\cite{NIU} &89.57	& 96.99 &+4.80 &2.10	&5.90	&14.90	&24.30	&20.10	&0.0658	\\
\midrule
Original   & 83.82 & 88.75 & 96.50 & 3.50 & 1.40 & 7.80 & 47.5 & 39.50 & - \\
\rowcolor{gray!5}SISA~\cite{bourtoule2021machine}    & 83.97 & 88.84 & 95.40 & 5.20 & 2.90 & 20.40 & 43.20 & 31.40 & 0.000 \\
\midrule
Original  & 81.20 & 100.00 & 100.00 & 0.00 & 0.00 & 0.00 & 64.00 & 0.00 & - \\
\rowcolor{gray!5}Fisher~\cite{golatkar2020eternal}  & 57.40 & 60.60 & -13.10 & 20.00 & 32.00 & 60.00 & 52.00 & 68.00 & 0.0002 \\
NTK~\cite{golatkar2020forgetting}   & 81.00 & 99.16 & +6.90 & 0.00 & 0.00 & 0.00 & 76.00 & 48.00 & 0.0001 \\
\bottomrule
\end{tabular}
}
\end{table*}

Regarding TA, most unlearning methods demonstrate consistent performance across different unlearning budgets, except for SSD, Amnesiac, and Fisher, which show significant variability.  In addition, the FA of PGU remains the closest to Retrain, while the FA of Boundary-E, Boundary-S, and Fisher rapidly decreases, indicating their ineffectiveness when unlearning a small number of data points.  
In Figure~\ref{fig:budget_one_cls}, L-CODEC effectively preserves model utility and maintains a lower FA when unlearning a small amount of data. However, as shown in Figure~\ref{fig:budget_all_cls}, the TA and FA of L-CODEC drop significantly as the unlearning budget increases, highlighting its inability with larger unlearning budgets.

\begin{remark}
\textbf{Remark 7.} \textit{Unlearning methods effective in specific scenarios and within certain budgets may be ineffective in others.}
\end{remark}

\section{Case Study}
\label{sec:case_study}

In Section~\ref{sec:evaluation}, we have evaluated unlearning methods in three different scenarios while neglecting the fact that certain data, which are highly representative or non-representative, can be easier or harder for the model to learn and memorize. Additionally, since unlearning methods are designed to erase the influence of specific data from a model, they could potentially be the defense strategy against data poisoning. Therefore, we further conduct three case studies: (1) Worst-case forgetting; (2) Best-case forgetting; and (3) Depoisoning.

\subsection{Case Study 1: Worst-case Forgetting}

We begin with the worst-case forgetting scenario. Instead of randomly selecting data to be unlearned, we select 1000 training samples with the lowest loss calculated with the original model $\theta_o$ as the forgetting dataset $\mathcal{D}_f$. These data are highly representative and are easily correctly classified by the model, even if the model has never seen them before. We hypothesize that most unlearning methods, especially those gradient-based ones,  would struggle to effectively unlearn in this case. \textbf{Table~\ref{tab:worst-case}} summarizes the results.

\begin{table*}[htbp]
\centering
\caption{Evaluation on CIFAR10 under the best-case forgetting scenario.}
\label{tab:best-case}
\resizebox{0.9\textwidth}{!}{%
\begin{tabular}{c|cc|cccccc|c}
\toprule
\rowcolor{gray!20}
\textbf{Method} & {\textbf{TA}} & {\textbf{RA}} & {\textbf{FA}} & {\textbf{correctness}} & {\textbf{confidence}} & {\textbf{entropy}} & {\textbf{m\_entropy}} & {\textbf{prob.}} & {\textbf{$\ell_2$}($\downarrow$)} \\
\midrule
Original & 93.24 & 100.00 & 100.00 & 0.30 & 17.60 & 100.00 & 91.80 & 3.00 & - \\
\hline
\rowcolor{blue!20}Retrain & 92.75 & 100.00 & 67.30 & 61.80 & 54.60 & 60.40 & 73.50 & 36.00 & 0.0000 \\
Amnesiac~\cite{heng2024selective} & 69.36 & 71.91 & -22.50 & 55.20 & 60.60 & 86.40 & 56.40 & 47.80  & 0.6698  \\
\rowcolor{gray!5}PGU~\cite{GradientProjectionUnlearning}      & 92.93 & 100.00 & +25.80 & 6.90 & 33.70 & 91.00 & 88.40 & 21.90 & 0.0209 \\
Unrolling~\cite{thudi2022unrolling} & 93.20 & 100.00 & +31.10 & 1.60 & 24.50 & 86.60 & 89.90 & 9.10 & 0.0463 \\
\rowcolor{gray!5}SCRUB~\cite{kurmanji2023towards} & 93.13 & 100.00 & +31.90& 0.80 & 22.00 & 89.00 & 85.00 & 11.20 & 0.0360 \\
$\ell_1$-Sparsity~\cite{jia2023model} & 92.68 & 100.00 & +30.20 & 2.50 & 50.00 & 81.50 & 85.70 & 17.10 & 0.0942 \\
\rowcolor{gray!5}First-order~\cite{warnecke2023machine} & 91.05 & 100.00 & +20.70 & 12.00 & 33.70 & 56.40 & 83.70 & 40.50 & 0.0208 \\
Second-order~\cite{warnecke2023machine} & 93.22 & 100.00 & +32.50 & 0.20 & 17.80 & 100.00 & 92.00 & 3.00 & 0.0208 \\ 
\rowcolor{gray!5}SSD~\cite{foster2024fast} & 65.40 & 70.60 & -26.70 & 59.40 & 52.60 & 46.90 & 61.90 & 76.70 & 0.0208 \\
\rowcolor{gray!5}Bad-T~\cite{chundawat2023can}  & 92.14 & 100.00 & +13.40 & 19.30 & 79.20 & 89.40 & 87.70 & 43.10 & 0.0224 \\
SalUn~\cite{fan2024salun}   & 92.17 & 100.00 & +8.60 & 24.10 & 91.70 & 91.70 & 94.80 & 80.40 & 0.0719 \\

\rowcolor{gray!5}L-CODEC~\cite{Mehta_2022_CVPR} & 73.97 & 80.00 & -23.20 & 55.90 & 60.20 & 65.30 & 60.50 & 71.00 & 0.0724 \\
Boundary-S~\cite{chen2023boundary}  & 92.72 & 100.00 & +30.80 & 1.90 & 37.30 & 74.70 & 83.20 & 15.70 & 0.0219 \\
\rowcolor{gray!5}Boundary-E~\cite{chen2023boundary}  & 92.77 & 100.00 &+31.00 & 1.70 & 37.30 & 74.90 & 79.70 & 15.70 & 0.0219 \\
FCS~\cite{FCS}	&92.07 & 99.65& +23.00 &9.70	&53.40	&75.40	&86.60	&37.90	&0.0626 \\
\rowcolor{gray!5}MSG~\cite{MSG}	&83.56	&88.00	&-5.90 &38.60	&46.20	&75.30	&62.10	&58.90	&0.0464	\\
CT~\cite{CT}	&65.81	&66.90	&-18.70 &61.30	&61.30	&48.60	&51.30	&51.20	&0.0464 \\
\rowcolor{gray!5}NIU~\cite{NIU} &85.08	&97.44	&+9.20 &23.50	&48.00	&67.00	&64.50	&58.80	&0.0658	\\
\midrule
Original    & 83.82 & 90.67 & 4.10 & 95.90 & 90.00 & 96.40 & 56.80 & 63.90 & - \\
\rowcolor{gray!5}SISA~\cite{bourtoule2021machine}    & 83.49 & 90.07 & 4.60 & 95.40 & 89.30 & 62.10 & 50.00 & 59.10 & 0.0000 \\
\midrule
Original  & 81.20 & 100.00 & 100.00 & 0.00 & 100.00 & 100.00 & 80.00 & 100.00 & 0.0002 \\
\rowcolor{gray!5}Fisher~\cite{golatkar2020eternal}  & 56.20 & 62.30 & 44.00 & 56.00 & 64.00 & 80.00 & 56.00 & 48.00 & 0.0002 \\
NTK~\cite{golatkar2020forgetting}   & 81.20 & 100.00 & 68.00 & 24.00 & 88.00 & 96.00 & 84.00 & 96.00 & 0.0001 \\
\bottomrule
\end{tabular}
}
\end{table*}

From the Retrain results, we observe that the model still achieves an FA of 93.10\% even without the forgetting data. Compared to the forgetting from all classes scenario in  Table~\ref{tab:random-all-cls}, the difference between Retrain from these two scenarios suggests that worst-case forgetting poses a substantial challenge for unlearning. Specifically, the FA of Retrain is 3.00\% higher in the worst-case scenario, and the MIA scores of Retrain are lower. This trend of higher FA and lower MIA scores is valid for all other unlearning methods in the evaluation,  regardless of whether they preserve model utility. Among them, SSD and MSG perform the best in balancing the model utility and unlearning efficacy. Specifically, SSD achieves 82.01\% in TA, 88.76\% in RA, and 66.20\% in FA, while MSG has 82.79\% in TA, 87.00\% in RA, and 78.10\% in FA. Both of them have relatively high MIA scores. However, the results in Table~\ref{tab:worst-case} suggest that all methods fail to effectively unlearn the forgetting data in the worst-case forgetting, underscoring the need for (1) the design of more effective unlearning methods and (2) a more comprehensive evaluation and analysis on the effectiveness of existing unlearning approaches.

\subsection{Case Study 2: Best-case Forgetting}
We next randomly select 1000 training data with the highest loss calculated with the original model $\theta_o$ as the $\mathcal{D}_f$. These data likely contribute minimally to model training and should, theoretically, be easier to unlearn. \textbf{Table~\ref{tab:best-case}} presents the results.

Compared to the results from the forgetting from all classes scenario in Table~\ref{tab:random-all-cls}, the FA of Retrain in Table~\ref{tab:best-case} is 22.8\% lower, and the MIA scores are significantly higher,
 suggesting that data that are inherently difficult for the original model to remember are easier to unlearn. In contrast to worst-case forgetting, best-case forgetting is characterized by lower FA and higher MIA scores across all evaluated unlearning methods. However, it can be observed from FA that most unlearning methods still cannot successfully unlearn the forgetting data. For example, PGU, Unrolling, SCRUB, $\ell_1$-Sparsity, First-order, Second-order, Bad-T, Boundary-S, Boundary-E, and FCS still exhibit FA values that are 25.80\%, 31.30\%, 31.90\%, 30.20\%, 20.70\%, 32.50\%, 13.40\%, 30.80\%, 31.00\%, and 23.00\% higher than that of Retrain, respectively. Particularly, SalUn, MSG, and NIU are among the methods that exhibit minimal FA discrepancies from Retrain, with +8.6\%, -5.9\%, and +9.2\%, respectively. Notably, only SalUn is able to maintain model utility. Additionally, Table~\ref{tab:best-case} implies that their MIA scores remain close to those
of the original model and are still relatively low, highlighting their deficiency when unlearning from all classes. 

Interestingly, the entropy-based MIA score on $\mathcal{D}_f$ of the original model reaches 100\%, indicating that high-loss forgetting data is already perceived as “unseen” by an attacker. After retraining without the forgetting dataset, the score drops to 60.4\%, with a similar trend for the m\_entropy-based MIA score. 

\subsection{Case Study 3: Depoisoning} 
At last, we evaluate whether an unlearning method can effectively remove the negative influence of poisoned data from a model and restore its original performance without retraining.

\textbf{Threat model.} We consider two types of poisoning attacks: (1) Label-flipping attacks, where an adversary aims to mislead the model prediction by flipping a portion of the labels in the training data. Label flipping can significantly reduce the performance of the original model $\theta_o$ on the source victim class. (2) Backdoor attacks~\cite{gu2019badnets}, where an adversary injects a trigger pattern, e.g., a small image patch, into a small portion of training data and modifies their labels towards a targeted incorrect label. Models trained on such poisoned datasets behave maliciously when encountering the trigger while appearing normal otherwise. Specifically, in label flipping, we flip labels between specific pairs of classes. For example, with a poison budget of 5000 samples, we randomly select 500 training samples from each of 10 classes in CIFAR10 and flip their labels as follows: $0\leftrightarrow 9$, $1\leftrightarrow 8$, $2\leftrightarrow 7$, $3\leftrightarrow 6$, $4\leftrightarrow 5$. In the backdoor attack, we randomly select 5000 training samples, add a 4$\times$4 patch to the bottom right corner of the images as a trigger pattern, and reassign their labels to class 0. For both depoisoning, we assume that model owners have identified the poisoned data and are applying machine unlearning techniques to mitigate the poisoning effects.

\begin{figure}[!t]
\centering
\includegraphics[width=\columnwidth]{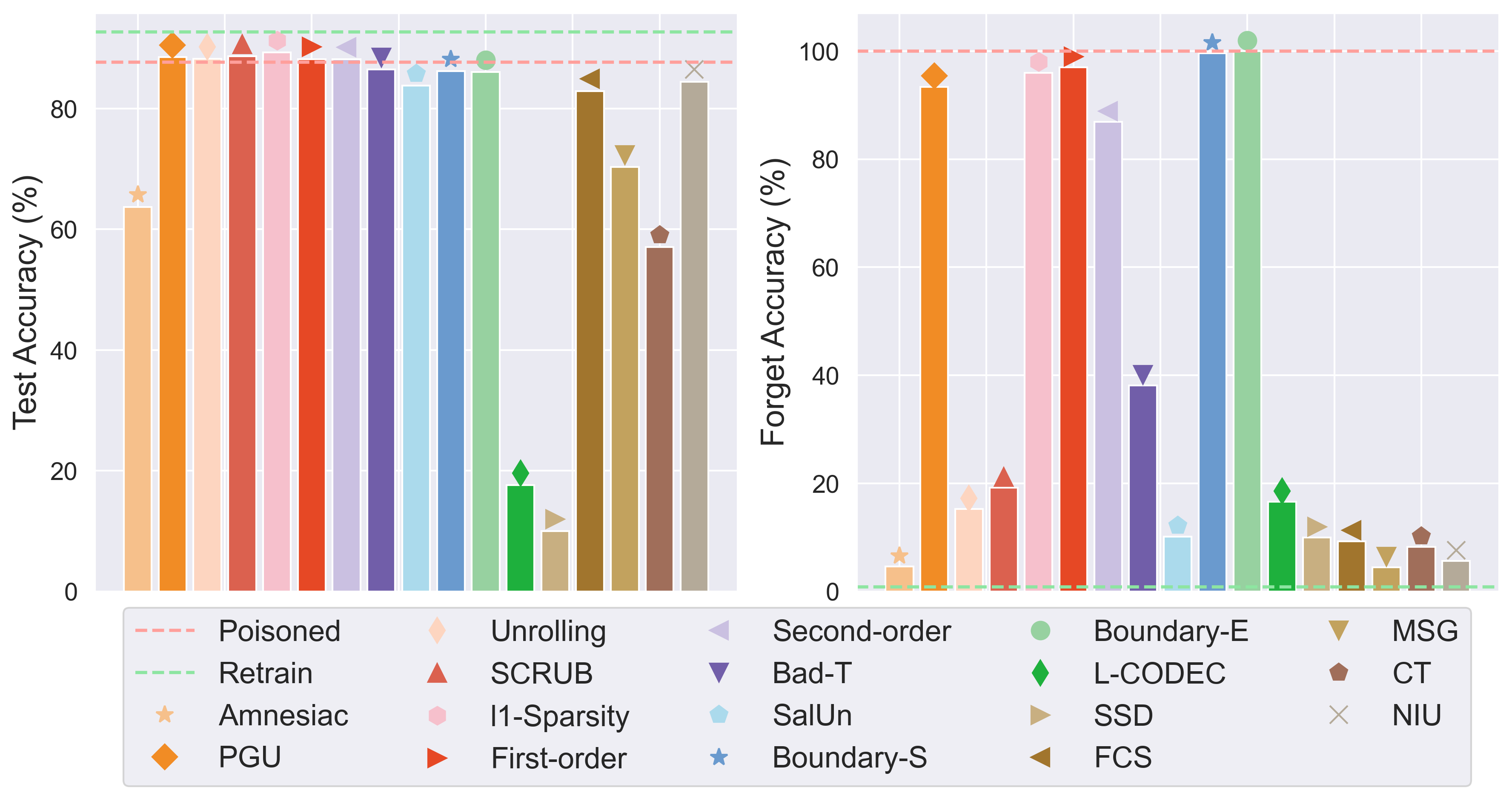}
\caption{TA and FA of different unlearning methods for ResNet18 trained on CIFAR10, when under the label flipping attack (2.5k poisoned data).}
\label{fig:depoison_2.5k}
\end{figure}

\begin{figure}[!t]
\centering
\includegraphics[width=\columnwidth]{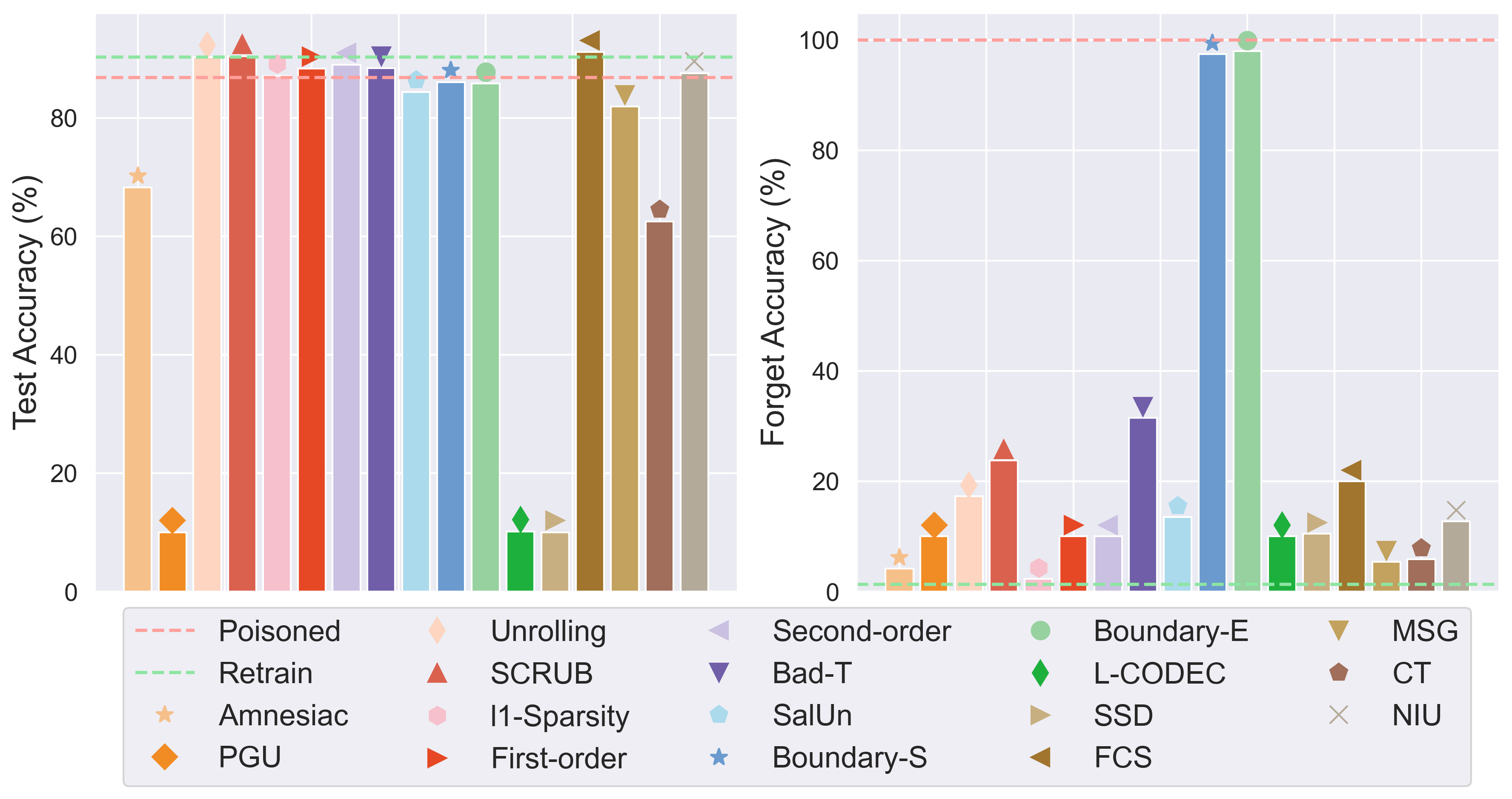}
\caption{TA and FA of different unlearning methods for ResNet18 trained on CIFAR10, when under the label flipping attack (5k poisoned data).}
\label{fig:depoison_5k}
\end{figure}

\textbf{Results on label flipping attacks.} \textbf{Figure~\ref{fig:depoison_2.5k}} and 
\textbf{Figure~\ref{fig:depoison_5k}} present the results, where the red and green dash lines denote the performances of the models trained with poison data and retrained without poison data, respectively. We make a similar observation as earlier that PGU, Unrolling, SCRUB, $\ell_1$-Sparsity, First-order, Second-order, and Bad-T can recover the test accuracy of the source victim class from poisoning. 
However, the FA, which can be interpreted as the attack success rate, of PGU, $l_1$-Sparsity, First-order, and Second-order remains high, indicating that while these methods improve test accuracy, they fail to unlearn the poisoned data. In contrast, Amnesiac, L-CODEC, SSD, MSG, and CT achieve low FA but at the cost of degraded TA. Among all methods, Unrolling, SCRUB, SalUn, FCS, and NIU appear to be the most effective depoisoning techniques for label-flipping attacks. 

A similar conclusion can be drawn from Figure~\ref{fig:depoison_5k}, but with notable differences. When the number of data to be unlearned increases from 2.5k to 5k, PGU fails to maintain TA. Additionally, Boundary-S and Boundary-E, which can preserve model utility under depoisoning, fail to unlearn the poisoned data, as their FA remains close to 100.00\%. Unrolling, SCRUB, $\ell_1$-Sparsity, First-order, Second-order, SalUn, and FCS continue to demonstrate effectiveness in achieving competitive TA while lowering FA. Among them, $l_1$-Sparsity stands out for best balancing model utility and unlearning efficacy. These findings highlight that existing unlearning methods lack consistent behavior when applied to the depoisoning task.

\begin{figure}[!t]
\centering
\includegraphics[width=\columnwidth]{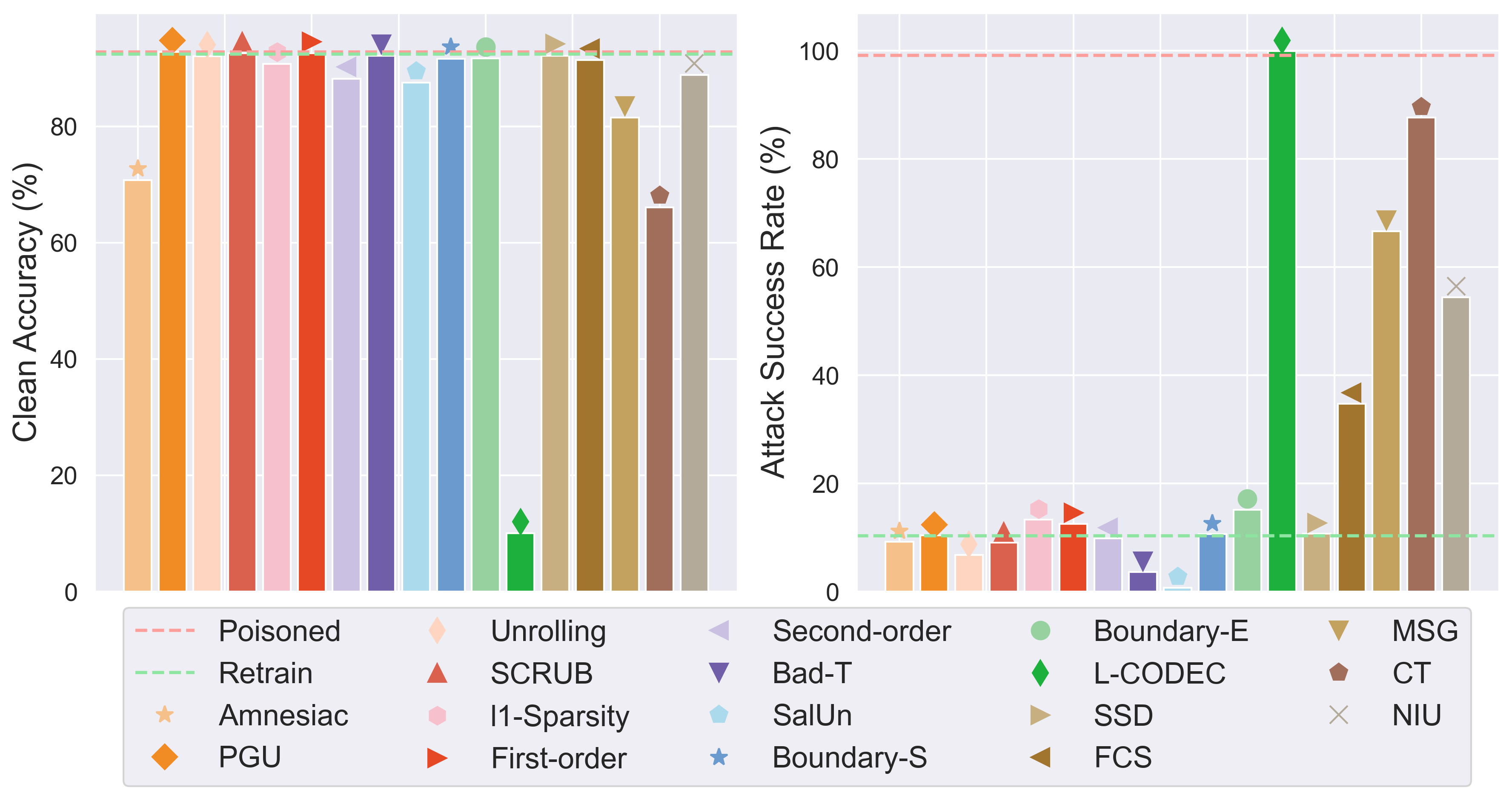}
\caption{Clean accuracy and attack success rate of different unlearning methods for ResNet18 trained on CIFAR10, when under the backdoor attack (5k poisoned data).}
\label{fig:modification_attack}
\end{figure}

\textbf{Results on backdoor attacks.} We then evaluate both the clean accuracy and attack success rate (ASR) after employing different unlearning methods to the backdoored CIFAR10 model. \textbf{Figure~\ref{fig:modification_attack}} illustrates the results. Depoisoning is considered successful if the model retains its clean accuracy while significantly reducing the ASR. Our results show that all unlearning methods effectively neutralize the backdoor effects, resulting in low ASR, while maintaining high accuracy, except for L-CODEC, FCS, MSG, CT, and NIU. L-CODEC causes the model to collapse into class 0, leading to only 10\% clean accuracy and nearly 100\% ASR, which severely compromises its utility. In contrast to depoisoning label-flipping attack, though FCS still maintains model's clean accuracy, the ASR can still achieve ~35\%. Similarly, while MSG, CT, and NIU can achieve low FA when depoisoning label-flipping attacks, the unlearned models still have high ASR. This indicates that neither method is effective for depoisoning in this context. Notably, the ASR of the Retrain method hovers around 10\% due to the presence of 10\% class 0 data in the test set, meaning the model correctly classifies data from class 0 and avoids misclassifying data from other classes with triggers.

These results show that despite the potential of machine unlearning in depoisoning, the effectiveness of depoisoning can significantly vary depending on the type and setting of the poisoning attack encountered.

\section{Conclusion and Future Work}
\label{sec:conclusion}

In this paper, we presented and evaluated MUBox, a uniform and comprehensive evaluation framework for machine unlearning in deep learning. MUBox currently integrates 23 state-of-the-art machine unlearning methods with 6 practical unlearning scenarios and
11 evaluation metrics. To the best of our knowledge, MUBox is the first platform that provides uniform, comprehensive, informative, and extensible evaluation of various unlearning methods. Leveraging MUBox, we conducted a systematic evaluation and clarified a number of open questions and challenges, uncovering various insights, including the design trade-offs between model utility and unlearning efficacy, the evaluation requirement of the worst-case forgetting, and the strength and limitation of existing methods.

While we aim to cover as broad a range of state-of-the-art unlearning methods as possible, MUBox does not exhaustively enumerate all methods. Our ongoing work will integrate more unlearning methods, model architectures, and datasets and 
expand to other domains, such as graph unlearning, federated unlearning, recommendation system unlearning, generative model unlearning, and other data modalities such as text and audio. 
We believe that MUBox will serve as a valuable benchmark to facilitate research in machine unlearning and highlight directions for further improvement.

\begin{acks}
The authors thank the partial support from Fordham AI Research Grant and Fordham Faculty Research Grant.
\end{acks}

\bibliographystyle{ACM-Reference-Format}
\bibliography{main.bbl}

\appendix

\vspace{-0.3cm}

\section*{Appendix}\label{appendix}

\subsection*{Reproducibility Study Summary}

Table~\ref{tab:statistics} presents 49 papers we analyzed in Section~\ref{sec:mu_stat_analysis}. 
From those papers whose codes are available, we incorporated 19 methods into MUBox. Note that several papers were not selected because they either do not fit our experiment settings~\cite{wu2020deltagrad,lin2023erm,ye2022learning,chien2024langevin,chien2024certified} or are built upon methods we had already included~\cite{gupta2021adaptive}.

\begin{table}[t]
\centering
\caption{Statistical analysis. This table includes 51 different methods from 49 papers. * denotes paper focusing on theoretical analysis.}
\label{tab:statistics}
\vspace{-0.3cm}
\resizebox{0.90\columnwidth}{!}{%
\begin{tabular}{c|cccc}
\toprule
\textbf{Method} & Has code? & README & Venue & Year \\
\midrule
\textbf{SISA}\cite{bourtoule2021machine}&\checkmark & \notcheckmark& IEEE S\&P&2020 \\
\textbf{First-order}\cite{warnecke2023machine}&\checkmark & \notcheckmark & NDSS & 2021\\
\textbf{Second-order}\cite{warnecke2023machine}&\checkmark & \notcheckmark & NDSS & 2021\\
\textbf{Unrolling}\cite{thudi2022unrolling}& \checkmark& \notcheckmark & Euro S\&P & 2022\\
\textbf{GKT}\cite{chundawat2023zero}&\checkmark & \notcheckmark& TIFS & 2023 \\
\cite{ginart2019making} & \ding{55} & - & NeurIPS & 2019 \\
\cite{gupta2021adaptive} & \checkmark & \checkmark & NeurIPS & 2021 \\
\cite{sekhari2021remember}*&\ding{55}&-&NeurIPS&2021\\

\cite{zhang2022prompt} & \ding{55} & - & NeurIPS & 2022 \\
\cite{tanno2022repairing}& \ding{55} & - & NeurIPS & 2022 \\
\cite{suriyakumar2022algorithms}& \ding{55} & - & NeurIPS & 2022 \\
\cite{chen2024fast} & \ding{55} & - & NeurIPS & 2023 \\
\textbf{SCRUB}~\cite{kurmanji2023towards} & \checkmark &\checkmark & NeurIPS & 2023 \\
\cite{liu2024certified}* & \ding{55} & - & NeurIPS & 2023 \\
\textbf{$\ell_1$-Sparsity}\cite{jia2023model} & \checkmark & \checkmark & NeurIPS & 2023 \\
\textbf{FCS}~\cite{FCS} & \checkmark & \notcheckmark & NeurIPS & 2023 \\
\textbf{MSG}~\cite{MSG} & \checkmark & \notcheckmark & NeurIPS & 2023 \\
\textbf{CT}~\cite{CT} & \checkmark & \notcheckmark & NeurIPS & 2023 \\
\textbf{NIU}~\cite{NIU} & \checkmark & \notcheckmark & NeurIPS & 2023 \\
\textbf{Langevin}\cite{chien2024langevin} & \checkmark & \checkmark & NeurIPS & 2024 \\
\textbf{PNSGD}~\cite{chien2024certified}& \checkmark & \checkmark & NeurIPS & 2024 \\
\textbf{SalUn}\cite{fan2024salun}&\checkmark & \checkmark& ICLR & 2024\\
\cite{shen2024label}& \ding{55}& -&ICLR & 2024\\
\cite{guo2019certified}&\checkmark&\checkmark&ICML&2020\\
\cite{wu2020deltagrad}& \checkmark&\checkmark & ICML& 2020 \\
\cite{chourasia2023forget}*& \ding{55}& - & ICML & 2023 \\
\textbf{Amnesaic}~\cite{graves2021amnesiac}& \checkmark &\notcheckmark & AAAI & 2021\\
\textbf{Bad-T}\cite{chundawat2023can}& \checkmark & \notcheckmark &AAAI & 2023\\
\cite{cha2024learning} & \ding{55} & - & AAAI & 2024 \\
\cite{wu2022puma} & \ding{55} & - & AAAI & 2022\\
\textbf{SSD}\cite{foster2024fast}&\checkmark & \checkmark& AAAI & 2024 \\
\cite{shibata2021learning}& \checkmark& \notcheckmark& IJCAI&2021 \\
\cite{yan2022arcane}&\ding{55} & -& IJCAI& 2022\\\
\textbf{Fisher}\cite{golatkar2020eternal}& \checkmark & \ding{55} & CVPR & 2020 \\
\cite{golatkar2021mixed}& \ding{55}&- & CVPR&2021 \\
\cite{kim2022efficient}&\ding{55} & -&CVPR & 2022\\
\textbf{L-CODEC}\cite{Mehta_2022_CVPR}&\checkmark & \checkmark&CVPR& 2022\\
\textbf{Boundary-S}\cite{chen2023boundary}& \checkmark & \notcheckmark & CVPR & 2023\\
\textbf{Boundary-E}\cite{chen2023boundary}& \checkmark & \notcheckmark & CVPR & 2023\\
\cite{lin2023erm}& \checkmark& \checkmark& CVPR&2023 \\
\cite{ye2022learning}& \checkmark&\checkmark & ECCV& 2022\\
\textbf{NTK}\cite{golatkar2020forgetting}& \checkmark& \ding{55}& ECCV&2020 \\
\cite{zhang2023closed}&\ding{55} & -& CIKM &2021 \\
\cite{lee2023undo} & \ding{55}& - & CIKM & 2023\\
\cite{pmlr-v134-ullah21a}*& \ding{55}& -& COLT& 2021\\
\cite{ghazi2023ticketed}* &\ding{55}&-&COLT&2023\\
\cite{neel2021descent}*&\ding{55}&-&ALT&2021\\
\cite{baumhauer2022machine}& \ding{55}&- & Machine Learning & 2022\\
\cite{zhang2022machine}& \ding{55}&- & ACM MM & 2022\\
\textbf{UNSIR}\cite{tarun2023fast}& \checkmark& \notcheckmark& TNNLS& 2023\\
\textbf{PGU}\cite{GradientProjectionUnlearning}& \checkmark& \notcheckmark & WACV &2024 \\
\bottomrule
\end{tabular}
}
\vspace{-0.3cm}
\end{table}

\subsection*{Key Ideas of Implemented Unlearning Methods}

Below, we briefly introduce the core concepts behind each unlearning method in the existing implementation. 
, i.e., $\theta_u = \mathcal{A}(\mathcal{D}_r)$, Although retraining from scratch is optimal for Machine Unlearning, it entails a large computational overhead, particularly for DNN training.


\textit{Exact unlearning}. Retrain involves retraining the model from scratch on the retaining dataset ($\mathcal{D}_r$) upon receiving unlearning requests. In SISA (Sharded, Isolated, Sliced, and Aggregated~\cite{bourtoule2021machine}), the original training dataset $\mathcal{D}$ is divided into $k$ disjoint shards, with $k$ sub-models independently trained on each shard. Upon unlearning requests, only the sub-model(s) trained on the shard containing the unlearning data need retraining. Additionally, data within each shard can be further divided into ``slices". Sub-models are incrementally trained on these slices. The weights are stored before including each new slice to track the influence of unlearned data points at a more fine-grained level, allowing retraining only from previous checkpoints. During inference, the global prediction is aggregated from each sub-model's predictions. 
(e.g., majority voting). 
\textbf{SISA}~\cite{bourtoule2021machine}(Sharded, Isolated, Sliced, and Aggregated), the original training dataset $\mathcal{D}$ is randomly partitioned into $k$ disjoint shards. $\mathcal{D}_1, \mathcal{D}_2, ..., \mathcal{D}_k$. Then, shard models $\theta^1, \theta^2,..., \theta^k$ are trained independently on each of these shards. Upon receiving an unlearning request, the model owner only needs to retrain the shard model that the unlearning data belongs to. Furthermore, when training each shard model on the shard dataset, SISA can further slice the data so that it only needs to retrain the corresponding shards from the previous checkpoints. During inference, the global prediction $k$ is simply aggregated from the predictions from each sub-model (e.g., majority voting). 

\textit{Gradient ascent}. Unrolling~\cite{thudi2022unrolling} expands a sequence of stochastic gradient descent (SGD) updates with a Taylor Series for gradient-based unlearning. When unlearning, it adds back the gradients of the unlearning data computed with respect to the initial weights to the final model weights. Amnesiac~\cite{heng2024selective} treats model training as a series of parameter updates to the initial model parameters. During the original training, it saves parameter updates from batches with data that need potential removal. When unlearning, it adds back the corresponding parameters updates to the model and retrains on the retaining dataset for a few epochs to restore model performance.

\textit{Influence function}. First-order~\cite{warnecke2023machine} uses a first-order Taylor Series of model $\theta_o$ to derive the gradient updates. Let $D_f = \{z_i = (x_i, y_i) \}_{i=1}^f$ be the data to be unlearned, and $\tilde{D}_f = \{\tilde{z}_i = z_i - \bm{\delta_i}\}_{i=1}^f$ be the corresponding unlearned counterparts, where $\bm{\delta}_i = (\delta_{x_i}, \delta_{y_i})$ is the unlearning modification for $(x_i, y_i)$. Then, it updates the model parameters as $\theta_u \leftarrow \theta_o - \tau\left(\sum_{\tilde{z}_i\in \tilde{Z}} \nabla \ell(\tilde{z}_i;\theta_o) - \sum_{z_i\in Z} \nabla \ell(z_i;\theta_o)\right)$, where $\tau$ is pre-defined unlearning rate and $\ell$ is the loss function. 
Second-order~\cite{warnecke2023machine} uses the inverse Hessian matrix of the second-order partial derivatives to update the original model's parameters. The unlearned model can be formalized as $\theta_u \leftarrow \theta_o - \textbf{\textit{H}}^{-1}_{\theta_o}\left(\sum_{\tilde{z}_i\in \tilde{Z}} \nabla \ell(\tilde{z}_i;\theta_o) - \sum_{z_i\in Z} \nabla \ell(z_i;\theta_o)\right)$. L-CODEC~\cite{Mehta_2022_CVPR} identifies a subset of model parameters with the most semantic overlap on forgetting data and applies Newton update only to this subset of parameters.

\textit{Teacher-student}. Both Bad-T~\cite{chundawat2023can} and SCRUB~\cite{kurmanji2023towards} leverage a teacher-student framework for unlearning. Bad-T uses a competent teacher (original model $\theta_o$) to preserve knowledge on the retaining dataset and an incompetent teacher (random model with the same structure) to destroy knowledge on the forgetting dataset. SCRUB optimizes a min-max problem, maximizing the Kullback-Leibler divergence (KLD) between the student model's output over the forgetting data and that of the teacher model, while minimizing the KLD over retaining data to maintain model utility.

\textit{Random labels}. Boundary-S~\cite{chen2023boundary} generates adversarial examples of forgetting data across the nearest decision boundary and assigns the adversarial labels to their corresponding forgetting data. It then fine-tunes the original model $\theta_o$ with forgetting data and their adversarial labels. Boundary-E~\cite{chen2023boundary} introduces an extra shadow class to the original model and assigns forgetting data to this shadow class for fine-tuning. SalUn~\cite{fan2024salun} selects the most salient weights for unlearning with respect to the forgetting data and updates only these weights when fine-tuning with randomly labeled forgetting data.

\textit{Fisher information}. Fisher~\cite{golatkar2020eternal} employs the Fisher information from the retaining dataset to unlearn specific samples, with Gaussian noise using the inverse of the Fisher Information Matrix as a covariance matrix to optimize the shifting effect. NTK~\cite{golatkar2020forgetting} approximates deep network activations as a linear function of weights and performs unlearning using the Fisher information matrix. SSD~\cite{foster2024fast} uses the Fisher information matrix of the training and forgetting data to select parameters that are disproportionately important to the forgetting dataset and induce forgetting by dampening these parameters proportional to their relative importance to the forgetting dataset concerning the wider training data.

\textit{Noise matrix}. 
UNSIR~\cite{tarun2023fast} proposes an Impair-Repair framework. It first learns an error-maximizing noise matrix for the class to be unlearned and manipulates the model weights to unlearn the targeted class of data (Impair). Then, it trains the model on the retaining dataset for a few epochs to regain the overall performance (Repair). Similarly, GKT~\cite{chundawat2023zero} generates error-maximizing noises to unlearn class(es) but replaces the repair step from UNSIR with error-minimizing noises that serve as retaining data to preserve the model utility and achieve zero-shot unlearning. Both UNSIR and GKT are class-wise unlearning approaches. 

\textit{Other representative approaches}. 
$l_1$-sparsity~\cite{jia2023model} first prunes the original model to obtain a sparsity model prior to unlearning and then fine-tunes it on the retaining dataset with sparsity regularization. PGU~\cite{GradientProjectionUnlearning} introduces an unlearning loss and uses SGD to update model parameters. It partitions the gradients into two orthogonal subspaces and updates weights in the direction orthogonal to the gradient subspaces. These weights are deemed unimportant for the retaining dataset to preserve model utility. FCS~\cite{FCS} employs a two-phase unlearning process: it first minimizes the KLD between the model's predictions on the forgetting set and a uniform distribution over the output classes and alternatively optimizes a contrastive loss between the model's outputs on the retaining and forgetting sets. Then, it fine-tunes the model on the retaining set to restore performance. MSG~\cite{MSG} collects gradients from the forgetting set via gradient ascent and from the retaining set via gradient descent. Using the collected gradient information, convolutional filter weights with the smallest absolute gradient values are re-initialized. The model is then retrained on the retaining set. CT~\cite{CT} transposes the weights of the convolutional layers in the model and fine-tunes the model on the retaining set. NIU~\cite{NIU} first re-initializes the parameters of the output layer to adjust the distribution of the final output while preserving the features learned by the model. It then randomly selects  $N$ layers from the network, repeatedly injecting noise into these layers and fine-tuning them. Finally, it fine-tunes all layers on the retaining set.

\subsection*{Additional Implementaion Details}
In our evaluation, we train the original ResNet18 and MobileViT on CIFAR10 and TinyImagenet using Nestrov SGD, respectively. We select the first 50 classes of the TinyImageNet. During training, we use a batch size of 256 and adopt a cosine learning rate scheduler for 200 epochs. We also apply basic augmentation techniques, including random cropping and horizontal flipping. 

Even though we strive to evaluate all methods using the exact same model architecture and dataset for a fair comparison, some minor inconsistencies still exist due to the scalability issue of some methods, limitations of computational resources, and some theoretical assumptions. Specifically, Fisher and NTK assume the model has been obtained by fine-tuning a pre-trained generic backbone. Therefore, we follow the following setting in experiments on CIFAR10; that is, we pre-train ResNet18 on CIFAR100 with the same training configuration stated above and then fine-tune the pretrained models on CIFAR10. Moreover, we are not able to make a consistent setting for NTK due to its lack of scalability both model size and dataset size. Instead, we follow the experimental setting in~\cite{golatkar2020forgetting}. In particular, we reduce the size of the ResNet18 to 40\% of the original model for both Fisher and NTK and obtain the small datasets following the same way in their experiments. In addition, due to the limitations of computational resources, we also reduce the size of the ResNet18 to 40\% of the original model for L-CODEC, while keeping other settings the same as those used for other unlearning methods in this paper. Nevertheless, this again underscores the importance of computational scalability and practicality when designing new unlearning algorithms.

Due to the limitation of available storage resources, the model unlearned by Amnesiac is only trained for 50 epochs, and other hyperparameters remain the same.

While we follow the hyperparameter settings provided by existing papers,
for Amnesiac, we finetune with 20 epochs after its unlearning procedure. For methods that use finetuning during their unlearning procedure. such as Unrolling, $\ell_1$-Sparsity and SalUn, we use 5, 20, 20 epochs, respectively.  For SISA, we use 5 shards and 1 slice. For fist-order, we use unlearning rate of 0.04 under forgetting from one class, worst-case and best-case forgetting scenarios, 0.08 under forgetting from all classes and class-wise forgetting scenarios, 0.00003 for depoisoning.

\subsection*{MIA Implementation Details}

MIA is implemented using the following five metrics: correctness~\cite{leino2020stolen}, confidence~\cite{shokri2017membership,yeom2018privacy,song2019privacy}, entropy~\cite{shokri2017membership}, modified entropy (m\_entropy)~\cite{song2021systematic}, and probability vector~\cite{shokri2017membership,song2021systematic}. In detail, we train the MIA predictor based on the model's correctness of predictions, the confidence level associated with the correct class, the output entropy of the model, the modified output entropy with the probability of the ground-truth class flipped, and the output probability vectors, respectively.

To train the MIA predictor, a balanced dataset is first sampled from the retaining dataset ($\mathcal{D}_r$) and the test dataset ($\mathcal{D}_t$). Then the five metrics are obtained from their output vectors, and used to train the MIA predictor, respectively. To evaluate the unlearning efficacy of MU methods, MIA-Efficacy is calculated by applying the trained MIA predictor to the unlearned model ($\theta_u$) on the forgetting dataset ($\mathcal{D}_f$). The goal is to determine how many data in $\mathcal{D}_f$ are classified correctly by the MIA model as "unseen" data with respect to $\theta_u$. Formally, 
\begin{equation}
\text{MIA-Efficacy} = \frac{TN}{|\mathcal{D}_f|},
\end{equation}

where $TN$ denotes the true negatives, i.e., the number of forgetting data predicted as ``unseen", and $|\mathcal{D}_f|$ is the size of the forgetting dataset.

\subsection*{Results on TinyImageNet}

\textbf{Table~\ref{tab:random-one-cls—tiny200}}, \textbf{Table~\ref{tab:random-all-cls-tiny500}}, and \textbf{Table~\ref{tab:class_tiny}} , \textbf{Table~\ref{tab:worst-case-tiny}}, and \textbf{Table~\ref{tab:best-case-tiny}} present the results under forgetting from one class, forgetting from all class, class-wise forgetting, worst-case forgetting, and best-case forgetting on TinyImageNet, respectively. Note that L-CODEC is excluded due to its extensive computation demand. In addition, since Fisher and NTK require that the model to be unlearned is pre-trained on other datasets, these two methods are not included either when evaluating on TinyImageNet.

While most results on TinyImageNet resemble those on CIFAR10, we make four observations regarding the scalability of existing unlearning methods across datasets and models. 
First, most methods, such as PGU, Unrolling, SCRUB, Boundary-S, and Boundary-E demonstrate good unlearning performance on CIFAR10 in class-wise forgetting. However, they fail to unlearn on TinyImageNet.
Second, the results of SSD in Table~\ref{tab:random-one-cls—tiny200}, Table~\ref{tab:class_tiny}, Table~\ref{tab:worst-case-tiny}, Table~\ref{tab:best-case-tiny} shows that it severely compromises the model utility on TinyImageNet while the method performs well on CIFAR10. Third, methods that use gradient information to unlearn, such as Unrolling, First-order, and Second-order, also demonstrate poor unlearning efficacy across all five unlearning scenarios. We hypothesize that this is due to the model being well-trained and overfitted to the training data. Thus their gradients are all close to 0, providing no effective information for unlearning. In their original papers, the evaluations are based on models that are not well-trained. However, as modern machine learning models are usually well-trained on their datasets, we suggest that unlearning methods should be evaluated in such a real-world setting. In addition, SISA can only achieve 18.40\% of test accuracy. This is because, in TinyImageNet, there are only 500 images in each class. After sharding the dataset, each submodel of SISA is only trained on an average of 100 data. Therefore, SISA cannot achieve a good classification accuracy for each class. These results suggest the poor generalization ability of existing unlearning methods in deep learning to other model architectures or datasets. 

\newpage

\begin{table*}[htbp]
\centering

\caption{Evaluation on TinyImageNet under the forgetting from one class scenario. 200 samples in class 0.}
\label{tab:random-one-cls—tiny200}
\resizebox{0.85\textwidth}{!}{%

\begin{tabular}{c|cc|cccccc}
\toprule
\rowcolor{gray!20}
\textbf{Method} & {\textbf{TA}($\uparrow$)} & {\textbf{RA}($\uparrow$)} & {\textbf{FA}($\downarrow$)} & {\textbf{correctness}($\uparrow$)} & {\textbf{confidence}($\uparrow$)} & {\textbf{entropy}($\uparrow$)} & {\textbf{m\_entropy}($\uparrow$)} & {\textbf{prob.}($\uparrow$)} \\
\midrule
Original & 59.88 & 100.00 & 100.00 & 0.00 & 0.00 & 1.00 & 100.00 & 0.00   \\
\hline
\rowcolor{blue!20}Retrain & 58.92 & 100.00 & 80.50 & 19.50 & 43.50 & 48.50 & 100.00 & 20.50  \\
Amnesiac~\cite{heng2024selective} & 31.24 & 36.44 & -23.50 & 43.00 & 42.50 & 32.50 & 100.00 & 51.50  \\
\rowcolor{gray!5}PGU~\cite{GradientProjectionUnlearning}      & 58.00 & 100.00 & -2.50 & 22.00 & 73.00 & 88.50 & 100.00 & 21.00  \\
Unrolling~\cite{thudi2022unrolling} & 59.52 & 100.00 & +19.50 & 0.00 & 3.00 & 6.00 & 100.00 & 0.00  \\ 
\rowcolor{gray!5}SCRUB~\cite{kurmanji2023towards} & 59.96 & 100.00 & +19.50 & 0.00 & 0.00 & 0.50 & 100.00 & 0.00   \\ 
$\ell_1$-Sparsity~\cite{jia2023model} & 59.36 & 96.77 & +8.00 & 1.50  & 3.00 & 12.00 & 100.00 & 2.50   \\ 
\rowcolor{gray!5}First-order~\cite{warnecke2023machine} & 57.88 & 100.00 & +6.70 & 12.80 & 79.20 & 82.60 & 100.00 & 100.00   \\ 
Second-order~\cite{warnecke2023machine} & 59.88 & 100.00 & +19.50 & 0.00 & 0.60 & 2.40 & 100.00 & 100.00 \\ 
\rowcolor{gray!5}SSD~\cite{foster2024fast} & 10.28 & 13.19 & -80.50 & 100.00 &100.00  & 44.00 & 100.00 & 55.00 \\ 
\rowcolor{gray!5}Bad-T~\cite{chundawat2023can}  & 58.84 & 100.00 & +14.00 & 5.50 & 100.00 & 14.50 & 100.00 & 58.00 \\ 
SalUn~\cite{fan2024salun}   & 57.16 & 100.00 & -74.00 & 93.50 & 98.00 & 45.00 & 100.00 & 84.00 \\ 
\rowcolor{gray!5}Boundary-S~\cite{chen2023boundary}  & 48.92 & 91.34 & -63.00 & 82.50 & 87.00 & 91.00 & 100.00 & 79.50  \\ 
Boundary-E~\cite{chen2023boundary}  & 48.80 & 91.33 & -63.00& 82.50 & 86.50 & 91.00 & 100.00 & 79.50  \\
\rowcolor{gray!5}FCS~\cite{FCS} &57.72 &	99.55&	+18.50 &1.00&	11.00&	27.50&	100.00&	2.50	\\
MSG~\cite{MSG} &20.76 &	21.34&	-29.50 &49.00&	48.50&	55.00&	100.00&	61.00	\\
\rowcolor{gray!5}CT~\cite{CT} &	43.68 &	47.80&	-8.00 & 27.50&	21.00&	31.50&	100.00&	26.50 \\	
NIU~\cite{NIU} & 49.04 &78.46&	-68.50 & 88.00&	100.00&	99.50	&100.00	&100.00	\\ 
\midrule
\rowcolor{gray!5}Original  & 18.40 & 7.37 & 1.00 & 1.00 & 1.00 & 26.50 & 100.00 & 76.50  \\ 
SISA~\cite{bourtoule2021machine}  & 18.36 & 7.68 & 1.05 & 1.50 & 2.50 & 25.50 & 100.00 & 77.20  \\ 
\bottomrule
\end{tabular}
}
\end{table*}

\begin{table*}[htbp]
\centering
\caption{Evaluation on TinyImageNet under the forgetting from all classes scenario. Forgetting 500 samples in all classes.}
\label{tab:random-all-cls-tiny500}
\resizebox{0.85\textwidth}{!}{%

\begin{tabular}{c|cc|cccccc}
\toprule
\rowcolor{gray!20}
\textbf{Method} & {\textbf{TA}($\uparrow$)} & {\textbf{RA}($\uparrow$)} & {\textbf{FA}($\downarrow$)} & {\textbf{correctness}($\uparrow$)} & {\textbf{confidence}($\uparrow$)} & {\textbf{entropy}($\uparrow$)} & {\textbf{m\_entropy}($\uparrow$)} & {\textbf{prob.}($\uparrow$)}  \\
\midrule
Original & 59.88 & 100.00 & 100.00 & 0.00 & 0.80 & 1.60 & 100.00 & 88.40  \\
\hline
\rowcolor{blue!20}Retrain & 58.48 & 100.00 & 55.60 & 44.40 & 62.80 & 62.20 & 100.00 & 90.80  \\
Amnesiac~\cite{heng2024selective} & 32.64 & 38.64 & -32.60 & 77.00 & 88.50 & 85.50 & 100.00 & 87.50  \\
\rowcolor{gray!5}PGU~\cite{GradientProjectionUnlearning}      & 8.12 & 11.33 & -47.20 & 8.40 & 7.00 & 37.60 & 100.00 & 85.20 \\
Unrolling~\cite{thudi2022unrolling} & 60.12 & 100.00 & +44.40 & 0.00 & 1.00 & 2.80 & 100.00 & 88.40  \\ 
\rowcolor{gray!5}SCRUB~\cite{kurmanji2023towards} & 60.40 & 100.00 & +44.40 & 0.00 & 0.40 & 2.00 & 100.00 & 88.40   \\ 
$\ell_1$-Sparsity~\cite{jia2023model} & 59.52 & 96.70 & +37.80 & 6.60 &11.10 & 29.20 & 100.00 & 88.40  \\ 
\rowcolor{gray!5}First-order~\cite{warnecke2023machine} & 58.16 & 100.00 & +44.40 & 0.40 & 2.00 & 6.80 & 100.00 & 88.60   \\ 
Second-order~\cite{warnecke2023machine} & 59.88 & 100.00 & +44.40 & 0.00 & 0.80 & 1.60 & 100.00 & 88.40  \\ 
\rowcolor{gray!5}SSD~\cite{foster2024fast} & 54.88 & 96.64 & +40.00 & 0.44 & 13.80 & 23.40 & 100.00 & 91.00  \\ 
\rowcolor{gray!5}Bad-T~\cite{chundawat2023can}  & 54.80 & 100.00 & +34.20 & 10.20 & 23.60 & 46.60 & 100.00 & 89.80  \\ 
SalUn~\cite{fan2024salun}   & 54.60 & 100.00 & -47.60& 92.00 & 96.80 & 53.20 & 100.00 & 87.80  \\ 
\rowcolor{gray!5}Boundary-S~\cite{chen2023boundary}  & 59.96 & 100.00 & +44.40 & 0.00 & 1.00 & 3.20 & 100.00  &88.40 \\ 
Boundary-E~\cite{chen2023boundary}  & 59.96 & 100.00 & +44.40 & 0.00 & 1.20 & 2.60 & 100.00 & 88.40  \\
\rowcolor{gray!5}FCS~\cite{FCS} &57.72	&99.56	&+41.60 &2.80	&25.80	&41.80	&100.00	&89.20	\\
MSG~\cite{MSG} &24.80	&26.69	&-30.20 &74.60	&74.60	&64.80	&100.00	&82.20	\\
\rowcolor{gray!5}CT~\cite{CT} &43.44	&48.13	&-8.40 &52.80	&48.20	&52.40	&100.00	&87.00	\\
NIU~\cite{NIU} &50.40	&0.78.78	&+22.00 &22.40	&64.20	&74.20	&100.00	&97.80	\\
\midrule
\rowcolor{gray!5}Original  & 18.40 & 7.66 & 72.60 & 72.60 & 100.00 & 72.80 & 100.00 & 78.80  \\ 
SISA~\cite{bourtoule2021machine}  & 19.44 & 7.49 & 39.80 & 39.80 & 50.80 & 60.40 & 100.00 & 77.20  \\ 
\bottomrule
\end{tabular}
}
\end{table*}

\begin{table*}[htbp]
\centering
\caption{Evaluation on TinyImageNet under the class-wise forgetting scenario. Forgetting all samples in class 0.}
\label{tab:class_tiny}
\resizebox{0.85\textwidth}{!}{%

\begin{tabular}{c|cc|cccccc}
\toprule
\rowcolor{gray!20}
\textbf{Method} & {\textbf{TA}($\uparrow$)} & {\textbf{RA}($\uparrow$)} & {\textbf{FA}($\downarrow$)} & {\textbf{correctness}($\uparrow$)} & {\textbf{confidence}($\uparrow$)} & {\textbf{entropy}($\uparrow$)} & {\textbf{m\_entropy}($\uparrow$)} & {\textbf{prob.}($\uparrow$)} \\
\midrule
Original & 59.88 & 100.00 & 100.00 & 0.00 & 0.60 & 2.40 & 100.00 & 100.00   \\
\hline
\rowcolor{blue!20}Retrain & 57.48 & 100.00 & 0.00 & 100.00 & 100.00 & 87.40 & 100.00 & 99.40  \\
Amnesiac~\cite{heng2024selective} & 24.60 & 27.94 & 0.00 & 100.00 & 100.00 & 26.40 & 98.20 & 96.80   \\
\rowcolor{gray!5}PGU~\cite{GradientProjectionUnlearning}      & 58.12 & 100.00 & +100.00 & 0.00& 27.40 & 40.80 & 100.00 & 100.00   \\
Unrolling~\cite{thudi2022unrolling} & 59.48 & 100.00 & 100.00 & 0.40 & 4.20 & 7.20 & 100.00 & 100.00  \\ 
\rowcolor{gray!5}SCRUB~\cite{kurmanji2023towards} & 59.76 & 100.00 & +100.00 & 0.00 & 0.20 & 0.60 & 100.00 & 100.00  \\ 
$\ell_1$-Sparsity~\cite{jia2023model} & 59.40 & 96.59 & +95.00 & 5.00 & 9.70 & 28.20 & 100.00 & 99.80  \\ 
\rowcolor{gray!5}First-order~\cite{warnecke2023machine} & 52.02 & 97.00 & +2.80 & 97.20 &100.00 & 88.40 & 92.20 & 94.40    \\ 
Second-order~\cite{warnecke2023machine} & 59.88 & 100.00 & 100.00 & 0.00 & 0.60 & 2.40 & 100.00 & 100.00  \\ 
\rowcolor{gray!5}SSD~\cite{foster2024fast} & 8.84 & 12.71 & 0.00 & 100.00 & 100.00 & 61.60 & 72.40 & 48.00  \\ 
\rowcolor{gray!5}Bad-T~\cite{chundawat2023can}  & 56.80 & 100.00 & +0.20 & 99.80 & 100.00 & 100.00 & 100.00 & 100.00  \\ 
SalUn~\cite{fan2024salun}   & 55.24 & 100.00 & +2.80 & 97.20 & 100.00 & 66.00 & 100.00 & 91.60  \\ 
\rowcolor{gray!5}Boundary-S~\cite{chen2023boundary}  & 43.88 & 80.38 & +4.40 & 95.60 & 96.40 & 83.80 & 53.20 & 93.20  \\ 
Boundary-E~\cite{chen2023boundary}  & 43.84 & 80.31 & +4.40 & 95.60 & 96.40 & 83.80 & 55.80 & 93.20  \\
\rowcolor{gray!5}FCS~\cite{FCS} &57.48	&99.54	&+76.60	&23.40 &23.40	&79.60	&0.922	&100.00	\\
MSG~\cite{MSG} &19.60	&21.28	&+1.20	&98.80 &97.00	&32.40	&98.80	&92.80	\\
\rowcolor{gray!5}CT~\cite{CT} &43.52	&47.57	&+65.60	&34.40 &76.60	&77.60	&95.00	&99.20	\\	
NIU~\cite{NIU} &49.04	&79.02	&0.00	&100.00 &100.00	&99.20	&100.00	&100.00	\\

\rowcolor{gray!5}GKT~\cite{chundawat2023zero}  & 3.40 & 3.39 & 0.00 & 0.00 & 0.00 & 34.20 & 1.00 & 77.40  \\
UNSIR~\cite{tarun2023fast} & 58.84 & 99.97 & +100.00 & 0.00 & 1.20 & 3.20 & 100.00 & 100.00  \\
\midrule
\rowcolor{gray!5}Original  & 18.40 & 7.69 & 1.40 & 1.40 & 0.80 & 25.80 & 94.00 & 84.00  \\ 
SISA~\cite{bourtoule2021machine}  & 17.88 & 7.69 & 0.00 & 0.00 & 0.00 & 26.60 & 85.60 & 84.00  \\ 
\bottomrule
\end{tabular}
}
\end{table*}

\begin{table*}[htbp]
\centering
\caption{Evaluation on TinyImageNet under worst-case forgetting scenario.}
\label{tab:worst-case-tiny}
\resizebox{0.85\textwidth}{!}{%

\begin{tabular}{c|cc|cccccc}
\toprule
\rowcolor{gray!20}
\textbf{Method} & {\textbf{TA}($\uparrow$)} & {\textbf{RA}($\uparrow$)} & {\textbf{FA}($\downarrow$)} & {\textbf{correctness}($\uparrow$)} & {\textbf{confidence}($\uparrow$)} & {\textbf{entropy}($\uparrow$)} & {\textbf{m\_entropy}($\uparrow$)} & {\textbf{prob.}($\uparrow$)} \\
\midrule
Original & 59.88 & 100.00 & 100.00 & 0.00 & 0.00 & 0.00 & 100.00 & 91.30   \\
\hline
\rowcolor{blue!20}Retrain & 59.00 & 100.00 & 93.60 & 6.40 & 16.70 & 17.50 & 100.00 & 91.70  \\
Amnesiac~\cite{heng2024selective} & 25.76 & 26.48 & -49.70 & 56.10 & 71.10 & 83.80 & 100.00 & 84.30  \\
\rowcolor{gray!5}PGU~\cite{GradientProjectionUnlearning}      & 59.44 & 100.00 & +6.40 & 0.00 & 0.00 & 0.00 & 100.00 & 91.30  \\
Unrolling~\cite{thudi2022unrolling} & 60.00 & 100.00 & +6.40 & 0.00 & 0.00 & 0.00 & 100.00 & 91.30  \\ 
\rowcolor{gray!5}SCRUB~\cite{kurmanji2023towards} & 59.12 & 100.00 & +6.40 & 0.00 & 0.00 & 0.00 & 0.00 & 91.30   \\ 
$\ell_1$-Sparsity~\cite{jia2023model} & 59.16 & 96.69 & +6.20 & 0.20 & 0.60 & 3.40 & 100.00 & 94.40   \\ 
\rowcolor{gray!5}First-order~\cite{warnecke2023machine} & 59.88 & 100.00 & +6.40 & 0.00 & 0.00 & 0.00 & 100.00 & 91.30   \\ 
Second-order~\cite{warnecke2023machine} & 59.88 & 100.00 & +6.40 & 0.00 & 0.00 & 0.00 & 100.00 & 91.30   \\ 
\rowcolor{gray!5}SSD~\cite{foster2024fast} & 59.88 & 100.00 & +6.40 & 0.00 & 0.00 & 0.00 & 100.00 & 91.30  \\ 
\rowcolor{gray!5}Bad-T~\cite{chundawat2023can}  & 46.20 & 95.50 & +2.70 & 3.70 & 15.40 & 24.70 & 100.00 & 91.30  \\ 
SalUn~\cite{fan2024salun}   & 53.16 & 98.88 & -77.20 & 83.60 & 96.40 & 74.00 & 100.00 & 91.70  \\ 
\rowcolor{gray!5}Boundary-S~\cite{chen2023boundary}  & 58.28 & 100.00 & +6.40 & 0.00 & 0.00 & 0.00 & 100.00 & 91.30   \\  
Boundary-E~\cite{chen2023boundary}  & 58.44 & 100.00 & +6.40 & 0.00 & 0.00 & 0.00 & 100.00 & 91.30   \\
\rowcolor{gray!5}FCS~\cite{FCS} & 57.36& 	99.44& 	+6.40&  0.00& 	0.50& 	1.20& 	100.00& 	91.30	 \\
MSG~\cite{MSG} & 21.12& 	21.35& 	-58.30&  64.70& 	64.60& 	54.30& 	100.00& 	83.10	 \\
\rowcolor{gray!5}CT~\cite{CT} & 43.56& 	46.99& 	-21.20&  27.60& 	22.50& 	41.70& 	100.00& 	89.90	 \\	
NIU~\cite{NIU} & 50.20& 	77.93& 	+1.90&  4.50& 	24.70& 	41.70& 	100.00& 	96.60	 \\
\midrule
\rowcolor{gray!5}Original  & 18.40 & 7.18 & 97.00 & 97.00 & 98.20 & 95.80 & 100.00 & 85.80  \\ 
SISA~\cite{bourtoule2021machine}  & 18.40 & 7.38 & 85.00 & 85.00 & 88.80 & 88.40 & 100.00 & 74.00  \\ 
\bottomrule
\end{tabular}
}
\end{table*}

\begin{table*}[htbp]
\centering
\caption{Evaluation on TinyImageNet under best-case forgetting scenario.}
\label{tab:best-case-tiny}
\resizebox{0.85\textwidth}{!}{%

\begin{tabular}{c|cc|cccccc}
\toprule
\rowcolor{gray!20}
\textbf{Method} & {\textbf{TA}($\uparrow$)} & {\textbf{RA}($\uparrow$)} & {\textbf{FA}($\downarrow$)} & {\textbf{correctness}($\uparrow$)} & {\textbf{confidence}($\uparrow$)} & {\textbf{entropy}($\uparrow$)} & {\textbf{m\_entropy}($\uparrow$)} & {\textbf{prob.}($\uparrow$)} \\
\midrule
Original & 59.88 & 100.00 & 100.00 & 0.30 & 0.142 & 0.803 & 100.00 & 87.20   \\
\hline
\rowcolor{blue!20}Retrain & 58.16 & 100.00 & 38.20 & 61.80 &  83.20& 80.01 & 100.00 & 89.40  \\
Amnesiac~\cite{heng2024selective} & 26.20 & 28.78 & -20.80 & 82.60 & 87.30 & 79.40 & 100.00 & 82.10  \\
\rowcolor{gray!5}PGU~\cite{GradientProjectionUnlearning}      & 45.00 & 82.94 & +11.80 & 50.00 & 38.40 & 52.40 & 100.00 & 81.20  \\
Unrolling~\cite{thudi2022unrolling} & 59.84 & 100.00 & +61.80 & 0.40 & 23.00 & 77.10 & 100.00 & 87.10  \\ 
\rowcolor{gray!5}SCRUB~\cite{kurmanji2023towards} & 60.08 & 100.00 & +61.80 & 0.30 & 18.20 & 63.30 & 100.00 & 87.20  \\ 
$\ell_1$-Sparsity~\cite{jia2023model} & 59.52 & 96.98 & +45.20 & 16.60 & 29.00 & 55.00 & 100.00 & 90.20   \\ 
\rowcolor{gray!5}First-order~\cite{warnecke2023machine} & 44.24 & 81.30 & +15.30 & 46.50 & 45.90 & 80.40 & 100.00 & 92.10   \\ 
Second-order~\cite{warnecke2023machine} & 59.92 & 100.00 & +61.50 & 0.30 & 14.20 & 78.90 & 100.00 & 87.20  \\ 
\rowcolor{gray!5}SSD~\cite{foster2024fast} & 16.20 & 21.30 & -24.50 & 86.30 & 78.40 & 37.60 & 100.00 & 63.00  \\ 
\rowcolor{gray!5}Bad-T~\cite{chundawat2023can}  & 51.84 & 98.00 &+27.40  & 34.40 & 60.00 & 85.30 & 100.00 & 92.50  \\ 
SalUn~\cite{fan2024salun}   & 54.20 & 100.00 & -33.50 & 95.30 & 98.60 & 60.50 & 100.00 & 89.80  \\ 
\rowcolor{gray!5}Boundary-S~\cite{chen2023boundary}  & 59.40 & 100.00 & +61.80 & 0.30 & 19.40 & 43.80 & 100.00 & 87.20  \\ 
Boundary-E~\cite{chen2023boundary}  & 59.52 & 100.00 &+61.80  & 0.30 & 19.50 & 45.40 & 100.00 & 87.20  \\
\rowcolor{gray!5}FCS~\cite{FCS} &58.00	&99.68	&+42.20 &19.60	&74.00	&84.10	&100.00	&90.40	 \\
MSG~\cite{MSG} &21.36	&22.82	&-23.60 &85.40	&44.30	&39.80	&100.00	&78.80	 \\
\rowcolor{gray!5}CT~\cite{CT} &43.60	&48.66	&-3.10 &64.90	&59.10	&59.10	&100.00	&87.40	 \\	
NIU~\cite{NIU} &50.36	&78.95	&+20.30 &41.50	&86.20	&88.70	&100.00	&99.10	 \\
\midrule
\rowcolor{gray!5}Original  & 18.40 & 7.76 & 60.20 & 60.20 & 87.20 & 65.60 & 100.00 & 77.20  \\ 
SISA~\cite{bourtoule2021machine}  & 18.08 & 7.84 & 27.60 & 27.60 & 27.60 & 64.60 & 100.00 & 74.80  \\ 
\bottomrule
\end{tabular}
}
\end{table*}

\end{document}